\documentclass[manuscript, screen]{jair}

\usepackage{subcaption}

\usepackage{mwe}
\usepackage{ragged2e}
\usepackage{array}
\usepackage[ruled,vlined]{algorithm2e}
\usepackage{algpseudocode}
\usepackage{varwidth}

\SetKwInput{KwInput}{Input}
\SetKwInput{KwOutput}{Output}
\SetKwInput{kwInit}{Init}

\usepackage{xcolor}

\usepackage{booktabs}

\usepackage{float}

\usepackage[utf8]{inputenc}
\usepackage{array}
\usepackage{makecell}

\setcopyright{cc}
\acmDOI{10.1613/jair.1.19264}

\JAIRAE{Pablo Samuel Castro}
\JAIRTrack{} 

\acmVolume{86}
\acmArticle{28}
\acmMonth{7}
\acmYear{2026}
\copyrightyear{2026}

\RequirePackage[
  datamodel=acmdatamodel,
  style=acmauthoryear,
  backend=biber,
  giveninits=true,
  uniquename=init
  ]{biblatex}

\begin{document}

\title{Hierarchical Reinforcement Learning with Optimal Level Synchronization Based on Flow-Based Deep Generative Model}
\author{JaeYoon Kim}
\authornote{Corresponding Author.}
\orcid{0000-0003-0898-5097}
\email{JaeYoon.Kim@student.uts.edu.au}



\author{Junyu Xuan}
\orcid{0000-0002-8367-6908}
\email{Junyu.Xuan@uts.edu.au}


\author{Christy Liang}
\orcid{0000-0001-7179-5208}
\email{Jie.Liang@uts.edu.au}


\author{Farookh Hussain}
\orcid{0000-0003-1513-8072}
\email{Farookh.Hussain@uts.edu.au}



\affiliation{%
  \institution{University of Technology Sydney}
  \city{Sydney}
  \state{New South Wales}
  \country{Australia}
}

\authorsaddresses{%
Authors' Contact Information:
JaeYoon Kim, \textsc{orcid:}
\href{https://orcid.org/0000-0003-0898-5097}{0000-0003-0898-5097},
\nolinkurl{JaeYoon.Kim@student.uts.edu.au};
Junyu Xuan, \textsc{orcid:}
\href{https://orcid.org/0000-0002-8367-6908}{0000-0002-8367-6908},
\nolinkurl{Junyu.Xuan@uts.edu.au};
Christy Liang, \textsc{orcid:}
\href{https://orcid.org/0000-0001-7179-5208}{0000-0001-7179-5208},
\nolinkurl{Jie.Liang@uts.edu.au};
Farookh Hussain, \textsc{orcid:}
\href{https://orcid.org/0000-0003-1513-8072}{0000-0003-1513-8072},
\nolinkurl{Farookh.Hussain@uts.edu.au}.
University of Technology Sydney, Sydney, New South Wales, Australia%
}



\renewcommand{\shortauthors}{Kim, Xuan, Liang \& Hussain}


\begin{abstract}
High-dimensional state and action spaces com-
bined with sparse reward structures in reinforcement learning
(RL) environments typically require advanced control architec-
tures. Hierarchical Reinforcement Learning (HRL) demonstrates
superior performance compared to atomic RL approaches in
these challenging scenarios. HRL can manage the complexity of
commands to achieve task objectives through its hierarchical
structure. One of the key challenges in HRL is efficiently
training each level’s policy with optimal data collection from
its experience. Off-policy correction is a critical technique
for facilitating sample-efficient off-policy training in HRL, as
it addresses the non-stationary issue of higher-level policy
training. However, existing methods typically employ indirect
probabilistic approaches that fail to accurately capture the
current capability of the lower-level policy. This mismatch
ultimately constrains the effectiveness of higher-level policy
training. In this paper, we propose a novel HRL model that
supports direct off-policy correction based on a Flow-based
Deep Generative Model (FDGM). This approach leverages the
inverse operation of FDGM to achieve goals aligned with the
current knowledge of the lower-level policy. Additionally, our
model addresses the limitations of FDGM to enable its effective
use in HRL. Through comparative experiments on benchmark
environments, our model demonstrates superior performance
over existing models.
    
\end{abstract}



\received{25 May 2025}
\received[accepted]{14 May 2026}

\maketitle

\section{Introduction}
Among many reinforcement learning (RL) algorithms, Hierarchical Reinforcement Learning (HRL) stands out by addressing the limitations of atomic RL algorithms such as Deep Q-Network (DQN) \cite{4}, Deep Deterministic Policy Gradient (DDPG) \cite{5}, Soft Actor-Critic (SAC) \cite{8}, and Twin Delayed Deep Deterministic Policy Gradient (TD3) \cite{9}. HRL demonstrates a strong capability to efficiently seek optimal solutions in high-dimensional state and action spaces, as well as sparse reward environments. This efficiency is achieved through temporal abstraction goals between the policies of HRL. Since HRL is composed of hierarchical policies, effective communication between higher-level and lower-level policies is crucial for its performance.

\begin{figure*}
 \centering
 \hspace*{-0.2cm} 
 \includegraphics[scale=0.25]{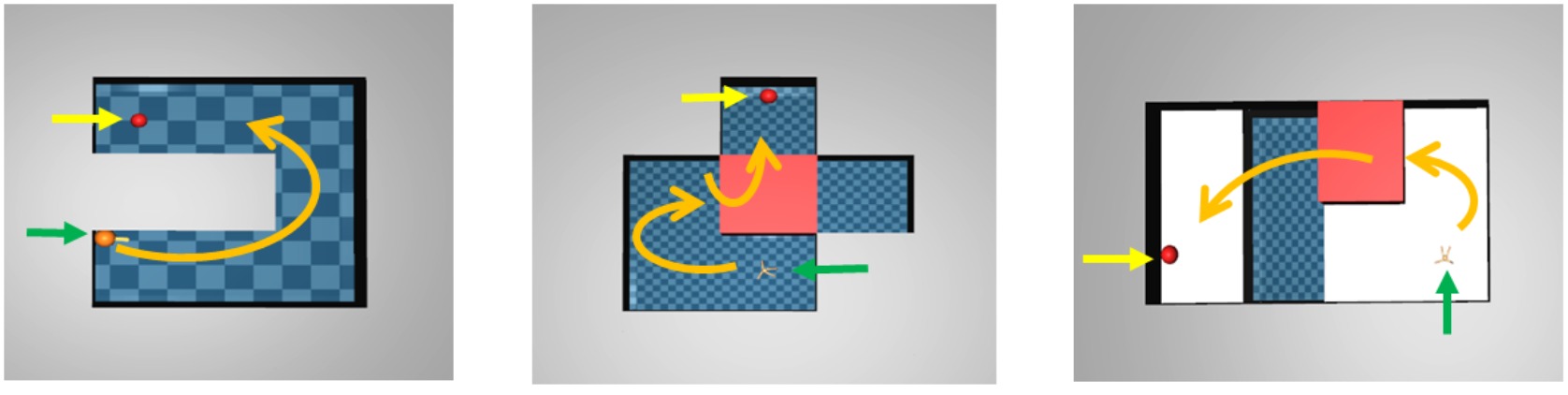}
 \vspace{-0.4cm}
 \caption{An example of several tasks in the Ant environment used in this research is shown above. The target location indicated by the yellow arrow serves as the reward point for the agent's approach. The agent, represented by the green arrow, must navigate to this target. The trained policy requires the agent to: (1) combine multiple action sequences while (2) interacting with a pink obstacle block in two out of the three above tasks. These tasks include: a series of directional movements (third from the right), clearing away an obstacle (second from the right), and constructing a bridge using an object (rightmost task). The higher-level policy guides its lower-level policy by providing abstract actions represented by the orange arrow, which specify the desired positions and orientation of the ant and its limbs to achieve the reward.}
 \label{fig:ant1}
\end{figure*}

In a two-level hierarchical structure, the higher-level policy focuses on abstract and generalized features, while the lower-level policy handles primitive actions in interaction with the environment. For instance, in the MuJoCo Ant environment, the hierarchical policy operates through coordinated temporal abstraction: (1) the higher-level policy generates macro-level movement directions visualized as arrows in Fig. \ref{fig:ant1}, which serve as subgoals, while (2) the lower-level policy executes corresponding primitive actions at each timestep to achieve these directional objectives.

In goal-conditioned HRL, a policy is conditioned on an additional input called a goal, which is provided by its higher-level policy. The communication between each HRL policy is crucial, and there are several approaches to managing this in an HRL algorithm. For example, an HRL algorithm can control a goal using an atomic RL algorithm \cite{18, 23, 58, 59} or various machine learning algorithms \cite{17, 21, 40}.

Meanwhile, the goal is also used for training its own policy. One of the key challenges in hierarchical RL, which is typically trained using a bottom-up layer-wise approach, is how to train each policy with optimal data collection for the task. Fig.  Off-policy correction for HRL, first introduced in HIRO \cite{18}, provides an effective solution to the optimal training challenge in HRL. As illustrated in Fig. \ref{fig:both_architecture}, HIRO's architecture implements this through a two-level policy framework with periodic goal correction.

HIRO combines off-policy training and correction to address the challenge of training a higher-level policy from the perspective of the current lower-level policy. Off-policy methods have been highlighted for their ability to overcome the local minima issues associated with on-policy methods. However, off-policy training also faces a non-stationary issue when training the higher-level policy of HRL due to its inherent algorithmic characteristics. 

The correction component requires careful consideration and examination. HIRO has attempted to address this issue through various indirect estimation methods, proposing an off-policy correction technique to mitigate the challenges of off-policy learning. This is achieved by relabeling past experiences, specifically the actions of the higher-level policy, as goals. 

However, the estimation process, which relies on an indirect probabilistic approach, has proven to be insufficient or imprecise due to its inherent limitations. The method's inability to accurately reflect the current knowledge of the lower-level policy stems from its reliance on indirect estimation.

Our research emphasizes the importance of direct estimation, which accurately reflects the current knowledge of the lower-level policy. This direct approach enhances the effectiveness of training the higher-level policy. In this study, we propose a novel HRL model with a direct off-policy correction strategy. This strategy leverages the current knowledge of the lower-level policy using a Flow-Based Deep Generative Model (FDGM), eliminating the need for superficial data derived from indirect estimations.

To reflect the current knowledge of the lower-level policy in the off-policy correction after training, we utilize the current internal parameters of its neural network (NN) to determine an exact goal for training the higher-level policy. By leveraging the inverse of the policy, we can directly compute the goal, effectively relabeling a previous goal of the higher-level policy.

When the result of an inverse operation of the lower-level policy is evaluated as a goal for the higher-level policy, reflecting the current knowledge of the lower-level policy, the estimation using the inverse model of the policy can serve as an accurate and effective off-policy correction method for HRL. In this context, determining the inverse value of a policy in a model-free HRL is defined as the inverse model of the policy.

The FDGM, capable of supporting the exact inverse value of a policy, is a promising candidate for defining an inverse model in HRL \cite{47, 48, 49, 50, 51, 55}. However, FDGM has inherent limitations, including a restriction between input and output dimensions, as well as a chronic issue—biased log-density estimation \cite{56, 57}. The former limitation makes it challenging to define an inverse model that uses a general method to support the flexible choice of goal dimensions. Meanwhile, ongoing research is addressing the latter issue in FDGM \cite{52, 53, 54}. Our research proposes an architecture inspired by \cite{39}, which is shown in Fig. \ref{fig:both_architecture}, to leverage the advantages of FDGM in HRL while mitigating its inherent limitations.

Our model is compared to HIRO in terms of the methodology for off-policy correction and the issue of biased log-density estimation in FDGM. Additionally, our model is compared to \textcolor{black}{Latent Space Policies (LSP)} \cite{23}, which uses FDGM for HRL shown in Fig. \ref{fig:both_architecture}, regarding the issue of the invertible bijective transformation function in FDGM mentioned above. \textcolor{black}{A comparative analysis between our proposed model and the reference models is presented in Table \ref{Comparison_model}}. 

In utilizing optimal off-policy correction to synchronize each level of HRL, several key questions arise. How can off-policy correction in HRL be effectively implemented using the current internal knowledge of a newly trained lower-level policy to optimize the training of its higher-level policy? Additionally, how can the limitations of FDGM be addressed when considering it for a lower-level policy, particularly when leveraging its inverse operation to relabel goals for off-policy correction?

Our experimental results on benchmark MuJoCo environments demonstrate that our approach outperforms HIRO and LSP in terms of both speed and performance accuracy. In summary, the contributions of this research are as follows:

\begin{itemize}
\item We propose a novel hierarchical RL model architecture that defines an inverse model for model-free HRL.
\item Our model adopts a direct method based on FDGM to identify an exact goal that reflects the current knowledge of a recently updated lower-level policy.
\item Despite utilizing FDGM in the lower-level policy, our model effectively addresses the limitations associated with FDGM in HRL.
\end{itemize}

The remainder of this paper is organized as follows. Section \ref{Preliminary Knowledge} introduces the preliminaries of HRL and off-policy correction. Section \ref{Related work} analyzes related research. Section \ref{Our model} describes our proposed HRL model using FDGM. Section \ref{Experiments} presents the experimental results comparing the performance of our HRL model with HIRO and LSP. Section \ref{Discussion} discusses the insights gained from the experiments. Finally, Section \ref{Conclusion} concludes this paper and outlines future work.

\section{Preliminary Knowledge}\label{Preliminary Knowledge} 

\subsection{ Hierarchical Reinforcement Learning}
\label{Preliminary Knowledge:Hierarchical reinforcement learning}
The structure of HRL is composed of several unit agents stacked hierarchically. A policy in goal-conditioned HRL typically has two inputs: the state of the environment $s_{t}$ and a goal $g_{t}$, which is a temporal abstraction provided by its higher-level policy $\mu_{hi}$. A unit agent is usually constructed based on an atomic RL framework, as it functions as a policy within a level of HRL. The lowest-level policy interacts with the environment by generating an action $a_{t} \sim \mu_{lo}(s_{t}, g_{t})$ based on the goal $g_{t}$ from its higher-level policy, where $g_{t} \sim \mu_{hi}$ is updated every \textit{c} time steps (\textit{c} is the horizon of $\mu_{hi}$) or determined by a fixed goal transition function \textit{h}($s_{t-1}, g_{t-1}, s_{t}$). Due to the hierarchical structure, the training period \textit{c} of a higher-level policy $\mu_{hi}$ is longer than that of its lower-level policy $\mu_{lo}$. The higher a policy is positioned in the HRL hierarchy, the more abstract its role becomes.

The environment responds to the action $a_{t}$ of the lowest-level policy with a reward $\textit{R}_{t}$ and a next state $s_{t+1}$, sampled from the reward function \textit{R}($s_{t}, a_{t}$) and the transition function \textit{f}($s_{t}, a_{t}$), respectively. The intrinsic reward $\textit{r}_{t}$ = \textit{r}($s_{t}, g_{t}, a_{t}, s_{t+1}$), where \textit{r} is a fixed parameterized reward function for the lower-level policy in HRL, is typically provided by the higher-level policy. Notably, the highest-level policy uses the environmental reward $\textit{R}_{t}$ instead of an intrinsic reward.

\textcolor{black}{Specifically, in Fig. \ref{fig:ant1}, the higher-level policy devises an abstract strategy (represented by the orange arrow), while the lower-level policy executes physical actions based on the goal from its higher-level policy and the current state. The model architectures, such as the one depicted in Fig. \ref{fig:both_architecture}, illustrate the outlined structure of HRL.} The key notations used in this research are summarized in Table \ref{notation}.

\subsection{Off-Policy Correction}
\label{Preliminary Knowledge:Off-policy correction}
It is crucial to devise an optimal training solution to enhance the performance of HRL. Off-policy methods in HRL have been proposed to address the local minimum issue associated with on-policy methods. In off-policy policy gradient methods, experience replay involves storing sampled trajectory data from past episodes in a replay buffer, which significantly improves sample efficiency. This approach enables better exploration by collecting samples using a behavior policy that differs from the target policy. The objective function of the off-policy policy gradient is defined as follows:
\begin{equation}
J(\theta) = \mathbb{E}_{S \sim d^{\beta}}\left[\sum_{a\in{A}} Q^{\pi}(s,a)\pi^{\theta}(a|s)\right]
\end{equation}
where $\textit{d}^{\beta}(s)$ is the stationary distribution of the behaviour policy $\beta(a|s)$,  $\pi^{\theta}(a|s)$ is the target policy and  $\textit{Q}^{\pi}(s,a)$ is the action-value function estimated with respect to the target policy $\pi^{\theta}(a|s)$ \cite{2}. 

However, the off-policy method in goal-conditioned HRL faces a non-stationary issue due to the bottom-up training approach in off-policy RL. Previously used outputs, such as goals from the higher-level policy, may no longer be suitable for training, as they do not reflect the updated parameter values $\theta$ of the newly trained lower-level policy. Consider the operation of a lower-level policy $\mu_{lo}$ in HRL as follows:
\begin{equation}
a_{t} \sim \mu_{lo}^{\theta} (s_{t}, g_{t}), \text{ at time} ~t. 
\end{equation} After $\mu_{lo}^{\theta}$ is trained for \textit{c} steps,
\begin{equation}
\begin{aligned}
a_{t+c} \sim \mu_{lo}^{\theta'} (s_{t+c}, g_{t})& \text{, assume $s_{t+c} = s_{t}$ }
\\
a_t \neq a_{t+c}{}\, \text{or} & \,a_t = a_{t+c}
\end{aligned}
\end{equation}
where $a_{t+c}$ and  $s_{t+c}$ are the action of $\mu_{lo}^{\theta'}$ and the state of the environment after \textit{c} steps, respectively. 
The action $a_{t+c}$ may differ from $a_{t}$ due to changes in the parameter values of $\mu_{lo}^{\theta}$ and $\mu_{lo}^{\theta'}$.

To address this issue, an off-policy correction is proposed in HIRO \cite{18} through an indirect estimation strategy.

\section{Related Work}\label{Related work}
\subsection{Hierarchical RL}\label{Hierarchical RL}
A large body of research has focused on developing optimal methods to connect adjacent level policies in the HRL framework from an action-oriented perspective. In the option-critic architecture, the policy over options selects an option to follow its intra-policy, which continues to operate until the condition for option termination is met \cite{12}. The Strategic Attentive Writer, with its HRL structure, offers advantages in sequential decision-making domains due to its use of macro-actions, which are partially organized based on environmental information \cite{14}. FeUdal networks for HRL identify semantically meaningful sub-goals in an environment, which are then pursued by different policies of the manager \cite{13}.

\begin{figure*}
  \vspace{-1.3cm}
  \centering
  \hspace*{0.5cm} 
  \includegraphics[scale=0.25]{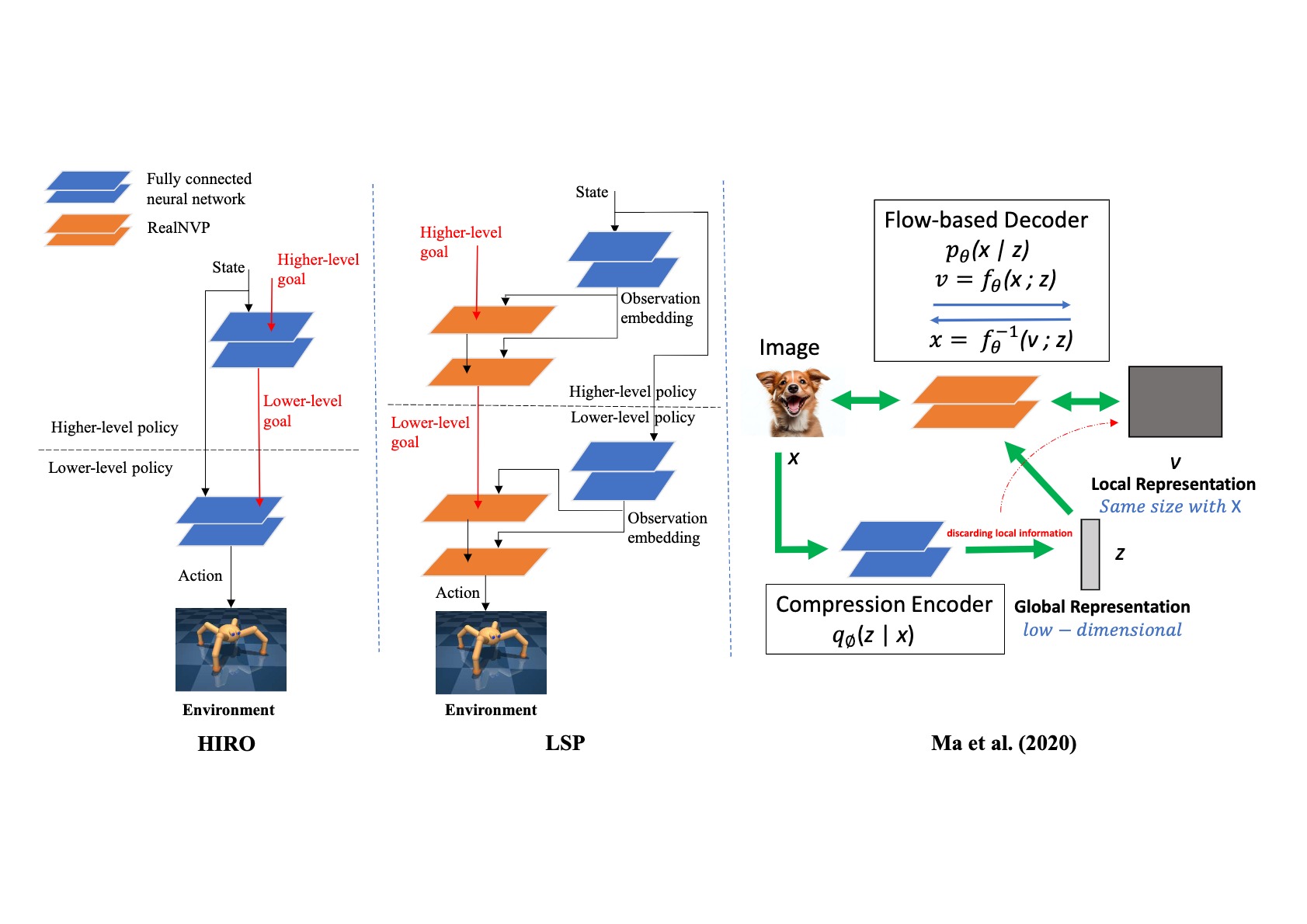}  
  \vspace{-1.8cm}
  \caption{The model architecture of HIRO (left), LSP (center) and Ma et al. (right)}
  \vspace{-0.8cm}
  \label{fig:both_architecture}
\end{figure*}

\cite{17} proposes an efficient and general method for discovering sub-goals using two unsupervised learning techniques: anomaly outlier detection and K-means clustering. The Hierarchical Actor-Critic incorporates DDPG with an actor-critic network and Hindsight Experience Replay (HER) \cite{10} at every policy level \cite{19}. \cite{20} introduces a nested and goal-conditioned HRL framework capable of dividing a task into several sub-tasks using two classes of hindsight transitions. \cite{16} develops a method to leverage representation learning through sub-optimality, defined as the difference between the value function of an optimal HRL that does not utilize representation learning and the value function of an HRL that does utilize representation learning.

\textcolor{black}{However, the aforementioned studies do not address the optimal training for level synchronization between adjacent level policies.} The main focus of HIRO is to develop an effective training method for each HRL policy, addressing the dynamics of training between adjacent level policies to ensure their synchronization. To identify an optimal goal $g_{t}\in{R^{d_{s}}}$ for off-policy correction in HRL, HIRO generates 10 candidate goals. These include eight candidates sampled randomly from a Gaussian distribution centered at the state difference $s_{t+c}- s_{t}$ (where ‘c’ is the horizon of the higher-level policy), the original goal $g_t$, and a goal equivalent to the state difference $s_{t+c}- s_{t}$. These candidates are evaluated based on the induced log probability of the lower-level policy:
\begin{equation}
\begin{aligned}
&\log\mu_{lo} (a_{t:t+c-1}|s_{t:t+c-1},\Tilde{g} _{t:t+c-1} )\\
\propto& -\frac{1}{2} \sum_{i=t}^{t+c-1} \|a_{i}- \mu_{lo} (s_{i},\Tilde{g}_{i}) \|_{2}^{2} + \text{const}
\end{aligned}
\end{equation} where $\mu_{lo}$ is the lower-level policy, $x_{t:t+c-1}$ represents the sequence $x_t,...,x_{t:t+c-1}$ collected by the higher-level policy, $a_t$ is the action of $\mu_{lo}$ and $\Tilde{g}_{t}$ is the relabeled goal $g_t$. 

\textcolor{black}{After identifying the best candidate, this goal is ultimately used to train the higher-level policy. The research also explores three additional methods, detailed in its Appendix. However, it does not propose a suitable method for off-policy correction in model-free HRL, as it relies on indirect estimation. Our research extends the work initiated by HIRO on off-policy correction, which is a central focus of our study.}

\textcolor{black}{LSP} proposes that each policy in HRL can be trained in a bottom-up, layerwise manner using a latent variable incorporated into the maximum entropy objective. To ensure that higher layers retain full expressivity, the latent variable and action must be invertible. This is achieved by leveraging \textcolor{black}{a FDGM}. In Fig. \ref{fig:both_architecture}, LSP demonstrates how a latent goal from the higher-level policy and an observation from the environment are provided to the lower-level policy. Two invertible coupling layers (orange), representing the FDGM in LSP, receive the latent goal from the higher-level policy and condition it on an observation embedding. The observation embedding, implemented with two fully connected layers (green), adjusts the dimension of the observation. \textcolor{black}{However, this approach is not suitable for off-policy correction in HRL, as it cannot support a flexible goal space due to the intrinsic limitations of FDGM, particularly the constraints of its invertible bijective transformation function.}

For our research, we adopt the use of FDGM in HRL, similar to \textcolor{black}{LSP}, while addressing its limitations.

\subsection{Deep Generative Models}\label{Deep generative models}
We have explored various deep generative models for use as an inverse model in HRL. After careful consideration, we concluded that employing a FDGM in our model ensures its validity as an inverse model in HRL.
\paragraph{\textbf{GAN and VAE}}
\cite{24} introduces an innovative deep generative model, Generative Adversarial Networks (GAN), which consists of two models: a discriminator and a generator. A Variational Autoencoder (VAE) \cite{26} employs a probabilistic encoder to generate a latent variable for a decoder, addressing the limitations of the vanilla Autoencoder \cite{27}. Numerous variants of these two generative models have been developed due to their robust unsupervised learning frameworks \cite{41, 42, 43, 36, 44, 45, 46}. However, GAN suffers from a significant drawback: training divergence caused by the difficulty in finding a Nash equilibrium \cite{25}. While VAE allows precise control over the latent variable, it often produces blurry outputs due to imperfect reconstruction, making it challenging to train the decoder effectively. Given these inherent disadvantages of GAN and VAE, we have opted to avoid their use in our research.

\paragraph{\textbf{Normalizing Flow}} 
A simple distribution can be transformed into a complex distribution by repeatedly applying an invertible mapping function. The change of variable theorem enables this transformation from one variable to another, resulting in the final distribution of the target variable as follows.

Suppose a probability density function $z \sim p(z)$ describes a random variable $z$. If there exists an invertible bijective transformation function $f$ between a new variable $x$ and $z$, such that $x = f(z)$ and $ z = f^{-1}(x)$, then the transformation is well-defined. Extending this to the multivariate case, if the change of variable theorem is applied to $\textbf{z}$ and a final target variable $\textbf{x}$ through successive distributions $p(\textbf{z})$ and successive transformation function $f(\textbf{z)}$, we have:
\begin{equation}
\begin{aligned}
\textbf{z}_{i-1} \sim\; p_{i-1}(\textbf{z}_{i-1}), ~  
\textbf{z}_{i} = f_{i}(\textbf{z}_{i-1}), \text{thus}\;
\textbf{z}_{i-1} = f_{i}^{-1}(\textbf{z}_{i}). 
\end{aligned}
\end{equation}
Then
\begin{equation}
\begin{aligned}
p_{i}(\textbf{z}_{i}) =& p_{i-1}(f_{i}^{-1}(\textbf{z}_{i}))\left| det  \frac{df_{i}^{-1}}{d\textbf{z}_{i}}   \right|
\\=& p_{i-1}(\textbf{z}_{i-1})\left| det  \frac{df_{i}}{d\textbf{z}_{i-1}}   \right|^{-1}, \text{thus}\\
\end{aligned}
\end{equation}

\begin{equation}
\begin{aligned}
\log{p_{i}(\textbf{z}_{i})} =& \log{p_{i-1}(\textbf{z}_{i-1}) - \log{\left|  det  \frac{df_{i}}{d\textbf{z}_{i-1}}    \right|}}
\end{aligned}
\end{equation}
where $det \frac{\partial f_{i}}{\partial \textbf{z}_{i}}$ is the Jacobian determinant of the function $f$.

Finally, the chain of K transformations of probability density function $f_{\textmd{i}}$—each of which is easily invertible and has a Jacobian determinant that can be efficiently computed—transforms the initial distribution $\textbf{z}_{0}$ into the final target variable $\textbf{x}$. This process is described as follows:
\begin{equation}
\begin{aligned}
\textbf{x} &= \textbf{z}_{\textmd{K}} = f_{\textmd{K}} \circ \circ \circ f_{2} \circ f_{1}(\textbf{z}_{0})\\
\log{p(\textbf{x})} &=\log{\pi_{\textmd{K}}(\textbf{z}_{\textmd{K}})} = \log{\pi_{0}(\textbf{z}_{0}) - \sum_{i=1}^{\textmd{K}} \log{\left| det  \frac{df_{i}}{d\textbf{z}_{i-1}}   \right|}}
\end{aligned}
\end{equation}

The flow arises from the path traced by the random variables $\textbf{z}_{i} = f_{i}(\textbf{z}_{i-1})$. A normalizing flow is defined as the full chain of transformations configured by the successive distributions $\pi_{i}$.

In our research, we focus on the advantages of normalizing flows, particularly their model flexibility and generation speed, despite their inherent drawbacks. The Real-valued Non-Volume Preserving algorithm (RealNVP) leverages normalizing flows, implemented using an invertible bijective transformation function. Each bijection, referred to as an affine coupling layer and denoted as $f: \textbf{x} \mapsto \textbf{y}$, decomposes the input dimension into two sections. The intrinsic transformation property of the affine coupling layer ensures that the input dimension remains unchanged, while alternately modifying the two split input sections in each coupling layer. This property enables straightforward inverse operations. Additionally, the Jacobian determinant is easily computed because the Jacobian matrix is lower triangular.

RealNVP employs a multi-scale architecture and batch normalization to enhance performance. To capture the local correlation structure of images, it utilizes two masked convolution methods: the spatial checkerboard pattern mask and the channel-wise mask \cite{28}. Non-linear Independent Components Estimation (NICE), a predecessor to RealNVP, uses an additive coupling layer that omits the scale term present in the affine coupling layer \cite{29}. Generative Flow with 1$\times$1 convolutions simplifies the architecture by addressing the reverse operation of channel ordering in NICE and RealNVP \cite{30}.

Several studies have attempted to address the chronic drawback of normalizing flows: biased log-density estimation \cite{37, 38}. In Fig. \ref{fig:both_architecture}, \cite{39} leverages the intrinsic characteristics of FDGM with an inductive bias based on its model architecture. By doing so, the research effectively utilizes the biased log-density estimation inherent to FDGM. The proposed model architecture combines a VAE, which extracts a global representation of an image, with \textcolor{black}{a FDGM} that relies on a local representation. The FDGM is conditioned on the global representation of the image provided by the VAE. As a result, an unbiased log-density estimation of the image can be achieved through the FDGM using the output of the VAE.

The compression encoder $q_{\phi}(z|x)$ in the VAE framework compresses the high-dimensional image $x$ into a low-dimensional global latent representation z, as follows:
\begin{equation}
q_{\phi}(z|x) = N(z; \mu(x), \sigma^{2}(x))
\end{equation}
where $\mu(\cdot)$ and $\sigma(\cdot)$ are neural networks that learn the mean and variance, respectively, and the variational posterior distribution $q_{\phi}(z|x)$ models the latent variable Z as a diagonal Gaussian.

To address the biased log-density estimation of FDGM in reconstructing the image $x$, $p_{\theta}(x|z)$ is considered using $z$. An invertible flow-based decoder $f_{\theta}(x;z)$ is employed, with the image $x$  as the main input and the global latent representation $z$   as the conditional input, to obtain a local representation $v$:
\begin{equation}
v = f_{\theta}(x;z)
\end{equation}
where $v$ follows a standard normal distribution, $v \sim N(0,I)$. Finally, the image x is reconstructed using the inverse function of the flow-based decoder:
\begin{equation}
x = f_{\theta}^{-1}(v;z).
\end{equation}

We adopt this architecture because it resolves two key challenges: the flexibility of the goal dimension between policies and the biased log-density estimation for the FDGM in our model.

\paragraph{ \textbf{Auto-Regressive Flow}}
An autoregressive flow provides tractable likelihoods because the probability of a full sequence can be expressed as the product of conditional probabilities of past observations. This property enables better density estimation compared to non-autoregressive flow-based models \cite{31, 32, 33, 34, 35}. However, the restoration speed of the original data, which is a consequence of its sequential operation, counterbalances this advantage.

\begin{table}[!t]
   \caption{Key Notations.}
   \begin{minipage}{\columnwidth}
     \begin{center}
       \begin{tabular}{lp{0.65\columnwidth}}
         \toprule
         Symbol & Meaning \\
         \hline
         \(t\) & action step\\
         \(\mu_{hi}\) & higher-level policy\\
         \(\mu_{lo}\) & lower-level policy\\
         \(\mu_{z}\) & the policy of  forward part\\
         \(\mu_{z-\text{state}}\) & the policy of  conditional part\\
         \(\mu_{\text{rnvp}}\) & the policy of  FDGM part\\
         \(\mu_{\text{rnvp}}^{-1}\) & the inverse of the policy of  FDGM part\\
         \(\Pi\) & the lower-level policy set composing of [$\mu_{z}$,$\,\,\mu_{z-\text{state}}$,$\,\,\mu_{\text{rnvp}}$]\\
         \(\textcolor{black}{L_{z}}\) & \textcolor{black}{the loss of forward part}\\
         \(\textcolor{black}{L_{z-\text{state}}}\) & \textcolor{black}{the loss of  conditional part}\\
         \(\textcolor{black}{L_{\text{rnvp}}}\) & \textcolor{black}{the loss of  FDGM part}\\
         \(\textcolor{black}{L_{\text{common}}}\) & \textcolor{black}{the averaged loss over the sum of loss of all three policies}\\
         \(g_{t}\:\textit{or}\:g\) & the action of  higher-level policy called a goal\\
         \(\Tilde{g_{t}}\) & a goal obtained through off-policy correction for training the higher-level policy $\mu_{hi}$\\
         \(a_{t}\) & the action of  lower-level policy\\
         \(R_{t}\) & the extrinsic reward of an environment \\
         \(r_{t}\) & the intrinsic reward of  lower-level policy\\         
         \(s_{t}\:\textit{or}\:s\) & the state of the environment\\
         \(a_{t,z}\) & the action of forward part\\
         \(a_{t,\text{rnvp}}\) & the action of  FDGM part of  inverse part\\
         \(a_{t,z-\text{state}}\) & the action of conditional part of inverse part\\
         \([g, a]\) & the concatenation of  ${g}_{1:d}$ and an action $a_{t,z-\text{state}}$\\
         \([a_{\text{rnvp}}, a_{z-\text{state}}]\) & the concatenation of an action $a_{t,\text{rnvp}}$ and an action $a_{t,z-\text{state}}$\\
         \(x_{t:t+c-1}\) & the sequence $x_{t}$,…,$x_{t+c-1}$\\
         \(c\) & the horizon of a higher-level policy\\
         \(D_{R}\) & replay buffer\\         
         \bottomrule
       \end{tabular}
     \end{center}
   \end{minipage}
   \label{notation}
 \end{table}

\begin{figure*}
  \vspace{-0.2cm}
  \centering
 \hspace*{-0.5cm}  
  \includegraphics[scale=0.45]
  {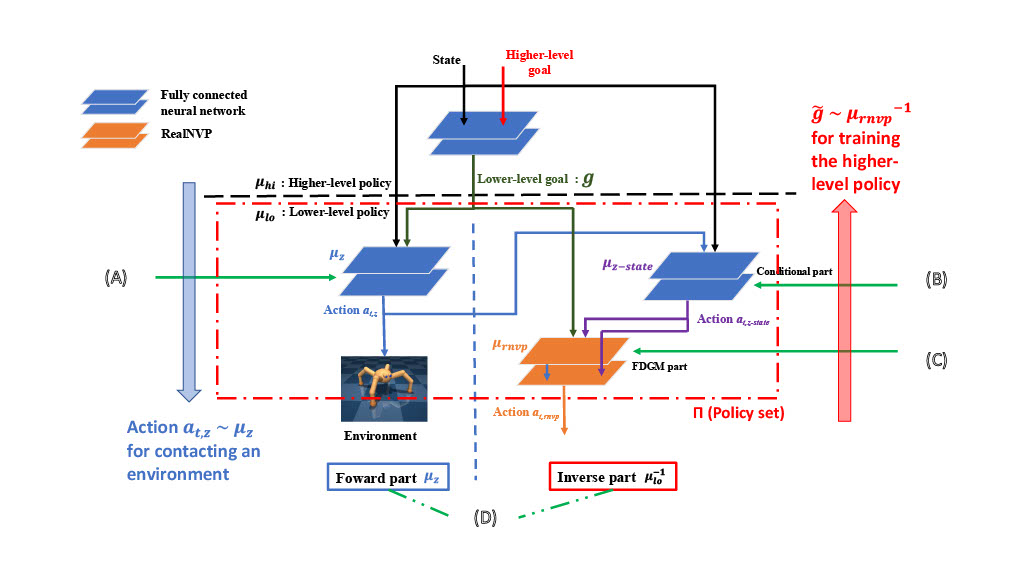}  
  \vspace{-1.3cm}  
  \caption{The architecture of our proposed model utilizes FDGM and incorporates the following key components: (A) A low-level policy from HIRO is employed exclusively for the forward part, which interacts with the environment. (B) The observation embedding of LSP, which functions as an encoder, is replaced by the low-level policy of HIRO, now treated as an independent policy. (C) The RealNVP component of LSP, a neural network integrated into a policy, is repurposed as an independent policy. (D) The combination of the forward and inverse parts is adopted from the model architecture of \cite{39}.}
  \label{fig:HRL_Our_Model}
\end{figure*}

\section{Our Model}\label{Our model}
The aim of our research is not to rely on indirect estimation, as in HIRO, but rather to employ direct estimation that accurately reflects the current knowledge of the lower-level policy. This approach enables us to find an optimal goal for level synchronization when training the higher-level policy. The most direct method is to determine a goal, $\Tilde{g_{t}}$, through the inverse operation of the lower-level policy, which is then used to train the higher-level policy $\mu_{hi}$. This is expressed as:
\begin{equation}
\Tilde{g_{t}} \sim \mu_{lo}^{-1}(a_{t})
\end{equation}
where $a_{t}$ is the action of lower-level policy, $\mu_{lo}$.

Given that FDGM supports invertible algorithms and fast generation speeds, we adopt an FDGM-based lower-level policy in our research to relabel goals for training the higher-level policy.

However, although FDGM is invertible, its bijective transformation property imposes constraints on goal dimensions, particularly when incorporating an inverse model into model-free HRL. This limitation stems from the requirement that the input and output dimensions of FDGM must be identical, making it difficult to design a model that accommodates flexible goal dimensions between policies in HRL. Additionally, FDGM suffers from the chronic drawback of biased log-density estimation.

To address these challenges, we define a lower-level policy set, denoted as $\Pi$, for the lower-level policy:
\begin{equation}
\label{eqn:lower-level_policy}
\Pi = [\mu_{z},\,\,\mu_{z-\text{state}},\,\,\mu_{\text{rnvp}}]
\end{equation}

We split the lower-level policy, $\mu_{lo}$, into two parts: a forward part composed of $\mu_{z}$, which interacts with the environment, and an inverse part, $\mu_{lo}^{-1}$, consisting of $\mu_{z-\text{state}}$ and $\mu_{\text{rnvp}}$ for the inverse model. The forward part is responsible for generating a real action in the environment. The action of the lower-level policy, denoted as $a_{t} \sim \mu_{lo}$, is determined by $\mu_{z}$ from $\Pi$ at the current state $s_{t}$. The inverse part includes two components: the $\mu_{\text{rnvp}}$ part, which uses the output of the conditional part, $\mu_{z-\text{state}}$, and a goal from the higher-level policy as inputs, and the conditional part, $\mu_{z-\text{state}}$, which supports the $\mu_{\text{rnvp}}$ part.

To obtain the local representation, the main inputs of the inverse part of our model are $s_{t}$ and $g_{t}$ for the FDGM part, and the output of the forward part, $a_{t,z}$, for the conditional part. However, the purpose of the inverse part is to find $\Tilde{g_{t}}$ using the inverse operation of the FDGM part. To achieve this, one of the inputs of the FDGM part, $s_{t}$, is transferred to the conditional part as an input. This design enhances performance and distinguishes our approach from the observation embedding of LSP, which functions solely as an encoder.

To ensure the exact inverse operation of our model with respect to the output of the corresponding level of the lower-level policy, it is critical to align the output of the inverse part, $a_{t,\text{rnvp}}$, with the output of the forward part, $a_{t,z}$. Essentially, both outputs, $a_{t,z}$ and $a_{t,\text{rnvp}}$, should share a common internal factor under the shared inputs—a goal $g_{t}$ and a state $s_{t}$—despite their differing dimensions. Therefore, $a_{t,z-\text{state}}$ of $\mu_{z-\text{state}}$ serves as the refined global representation.

Through $\Pi$, our model ensures the flexibility to choose the dimension of the goal at any level while addressing the issue of biased log-density estimation in FDGM. Since off-policy correction is not required at the highest level, all levels except the highest can adopt the two-part architecture of our model: a forward part and an inverse part.

Fig. \ref{fig:HRL_Our_Model} illustrates the full architecture of our model, which is based on a two-level hierarchical structure. Each component of the lower-level policy, $\mu_{lo}$, including the off-policy correction employed in our model, is explained in detail in the following section.

\begin{algorithm}[t]
  
  \caption{Find the goal, $\Tilde{g}$, using only $\mu_{rnvp}^{-1}$ without forward part (Normal version)}
  \KwInput{$a_{z-\text{state}}$ and $a_\text{rnvp}$}
  \KwOutput{$\Tilde{g}$}
  \kwInit{
    \begin{itemize}
      \item Set the layer dimension of all policies including $\mu_{hi}$ and the critic of $\mu_{rnvp}$  to  be the same except for the actor of $\mu_{rnvp}$ 
      \item Set the layer dimension of actor of $\mu_{rnvp}$ 
      \item Set the action dimension of $\mu_{z-\text{state}}$
    \end{itemize}
  }

  \While{$c$}{
        $a_{z-\text{state}}\;and \;a_\text{rnvp} \sim D_{R}$ \;
        
        $\Tilde{g} \sim \mu_\text{rnvp}^{-1}(a_{z-\text{state}}, a_\text{rnvp})$ \;
    }
  \label{alg:the_alg1}
\end{algorithm}

\begin{algorithm}[t]
  
  \caption{Find the goal, $\Tilde{g}$, using only $\mu_{rnvp}^{-1}$ with forward part (Variant version)}
  \KwInput{$s, g\;and\;a_\text{rnvp}$}
  \KwOutput{$\Tilde{g}$}
  \kwInit{
    \begin{itemize}
        \item Set the layer dimension of all policies (a forward part and an inverse part) inside $\Pi$  to be the same except for  $\mu_{hi}$ 
        \item Set the layer dimension of $\mu_{hi}$ 
        \item Set the action dimension of $\mu_{z-\text{state}}$
    \end{itemize}
  }

  \While{$c$}{
      $s, g\;and\;a_\text{rnvp} \sim D_{R}$ \;
      
      $a_{z}^{'} \sim \mu_{z}(s, g)$ \;
      
      $a_{z-\text{state}}^{'} \sim \mu_{z-\text{state}}(s, a_{z}^{'})$ \;
      
      $\Tilde{g} \sim \mu_\text{rnvp}^{-1}(a_{z-\text{state}}^{'}, a_\text{rnvp})$
    }
  \label{alg:the_alg2}
\end{algorithm}

\subsection{Forward Part $\mu_{z}$}\label{Forward part}
Similar to the lower-level policy in HRL, the action of the forward part interacts with either the environment or another lower-level policy. We model the forward part using the VAE framework proposed by \cite{39}, which extracts the global representation of a goal $g_{t}$ and a state $s_{t}$ in HRL as follows: \begin{equation}
a_{t,z} \sim \mu_{z}(s_{t}, g_{t}) \text{ at time } ~t
\end{equation}
where $\mu_{z}$ is the policy of the forward part, and $a_{t,z}$ is the action of $\mu_{z}$. The action $a_{t,z}$ serves as the global representation, which aids in finding the goal $\Tilde{g}_{t}$ in the inverse part for relabeling $g_{t} \sim D_{R}$.

The fixed dimension of the action $a_{t,z}$, determined by the given environment and distinct from the dimension of $g_{t}$, compensates for a potential limitation of the inverse part, which relies on an FDGM and requires dimension matching with $g_{t}$. Since the outputs of both parts, $a_{t,z}$ and $a_{t,\text{rnvp}}$,  are equivalent from the perspective of the higher-level policy, we can treat $a_{t,z}$—the output of the forward part—as a global representation corresponding to the output of the inverse part, $a_{t,\text{rnvp}}$.

\subsection{Inverse Part $\mu_{lo}^{-1}$}\label{Inverse part}
We aim to utilize the inverse operation of FDGM to compute a goal $\Tilde{g}_{t}$ that aligns with the updated lower-level policy, $\mu_{lo}$. Our model architecture is inspired by the approach of \cite{39}, which employs two distinct representations to balance the strengths and weaknesses of FDGM.

\subsubsection{Conditional Part $\mu_{z-\text{state}}$}\label{Conditional part}
While the conditional part serves a role similar to the observation embedding in LSP as a conditional input for FDGM, it differs in two key aspects. First, unlike the observation embedding in LSP, which is not trained as an encoder using a fully connected neural network, the conditional part operates as an independent policy to enhance performance. Second, $\mu_{z-\text{state}}$ takes both the state $s_{t}$ and action $a_{t,z}$ as inputs to process the global representation, whereas the observation embedding in LSP only takes the state $s_{t}$ to adjust its dimension. This is expressed as:\begin{equation}
a_{t,z-\text{state}} \sim \mu_{z-\text{state}} (s_{t}, a_{t,z}) \text{ at time} ~t 
\end{equation}
where $\mu_{z-\text{state}}$ is the policy of the conditional part and $a_{t,z}$ is the action of the forward part.

There are three reasons why we incorporate the conditional part. First, it provides a refined global representation $a_{t,z}$ while preserving the local representation $a_{t,\text{rnvp}}$ as a conditional input for FDGM. Second, to address the chronic weakness of the FDGM part—biased log-density estimation—we account for the influence of the conditional part, which includes the global representation of a goal $g_{t}$ and a state $s_{t}$.

The second reason is that controlling the dimension of the action $a_{t,z-\text{state}}$ enhances the effectiveness of $a_{t,z}$  in the FDGM part. Although $a_{t,z}$ is fixed by the environment, we aim to indirectly influence its dimension through $a_{t,z-\text{state}}$ to optimize its impact on the output of the FDGM part.

The third reason is to adjust the dimension of the state $s_{t}$ if it is greater or less than that of the goal $g_{t}$, as both are original inputs to the FDGM part. This adjustment helps achieve a more accurate $\Tilde{g_{t}}$ through $\mu_{lo}^{-1}$.

Finding the optimal dimension for $a_{t,z-\text{state}}$ involves a trade-off process that can be refined through experimentation.

\subsubsection{FDGM Part $\mu_{\text{rnvp}}$}\label{FDGM part}
The inputs to the FDGM part are a goal $g_{t}$, provided by the higher-level policy as the main input, and an action $a_{t,z-\text{state}}$, provided by the conditional part as the conditional input. This is expressed as: 
\textcolor{black}{\begin{equation}
a_{t,\text{rnvp}} \sim \mu_{\text{rnvp}} (g_{t}, a_{t,z-\text{state}}) \text{ at time} ~t
\end{equation}} where $\mu_{rnvp}$ is the policy of FDGM part and $a_{t,\text{rnvp}}$ is the action of $\mu_{rnvp}$. The principle of the FDGM part can be explained using the forward transformation property of RealNVP \cite{28} in a single layer:
\begin{equation}
\begin{aligned}
{a\_\text{rnvp}}_{1:d} &= {g}_{1:d}\\
{a\_\text{rnvp}}_{d+1:D} &= {g}_{d+1:D}{\odot} \exp(s({[g, a]})) + t({[g, a]})
\end{aligned}
\end{equation}
where ${\odot}$ represents the element-wise product. A goal ${g}_{t}$ is split into \textcolor{black}{a} ${g}_{1:d}$ and \textcolor{black}{a} ${g}_{d+1:D}$. The concatenation of ${g}_{1:d}$ and the action $a_{t,z-\text{state}}$ forms ${[g, a]}$. The outputs of a coupling layer, ${a\_\text{rnvp}}_{1:d}$ and ${a\_\text{rnvp}}_{d+1:D}$ are concatenated for the next layer. Furthermore, $s(\cdot)$ and $t(\cdot)$ are scale and translation functions that map $R^{d} \mapsto R^{D-d}$. These functions are implemented using a deep neural network, as the inverse function of each bijection and the Jacobian determinant do not require explicit computation of s and t. The inverse operation, along with the forward operation, can be calculated as follows: 
\begin{equation}\begin{split}
    \left\{
    \begin{array}{c}
        a\_\text{rnvp}_{1:d} = {g}_{1:d}\\
        a\_\text{rnvp}_{d+1:D} = {g}_{d+1:D}{\odot}
        \exp(s({[g, a]})) + t({[g, a]})
    \end{array}
    \right.
\end{split}\end{equation}
\begin{equation}\begin{split}
\Longleftrightarrow{
    \left\{
    \begin{array}{c}
        g_{1:d} = {a\_\text{rnvp}}_{1:d}\\
        g_{d+1:D} = ({a\_\text{rnvp}}_{d+1:D}-t([a_\text{rnvp}, a_{z-\text{state}}]))\\\;\;\;\;\;{\odot}
        \exp{(-s([a_\text{rnvp}, a_{z-\text{state}}]))}
    \end{array}
    \right.
}
\end{split}\end{equation}
Here, an action $a_{t,\text{rnvp}}$ is split into $a\_\text{rnvp}_{1:d}$ and $a\_\text{rnvp}_{d+1:D}$. The concatenation of $a_{t,\text{rnvp}}$ and $a_{t,z-\text{state}}$ forms \textcolor{black}{an} [$a_\text{rnvp}, a_\text{z-state}$]. The outputs of a coupling layer in the inverse operation are $g_{1:d}$ and $g_{d+1:D}$.

The dimension of $a_{t,\text{rnvp}}$ is determined by the dimension of $g_{t}$ due to the bijective transformation property of FDGM. The output of the FDGM part, $a_{t,\text{rnvp}}$, is not used to provide an action to the environment but rather for the off-policy correction of its higher-level policy after being stored in the replay buffer $D_{R}$.

\begin{figure}
  \vspace{-0.2cm}
  \centering
  \hspace*{0.4cm} 
  \includegraphics[scale=0.15]{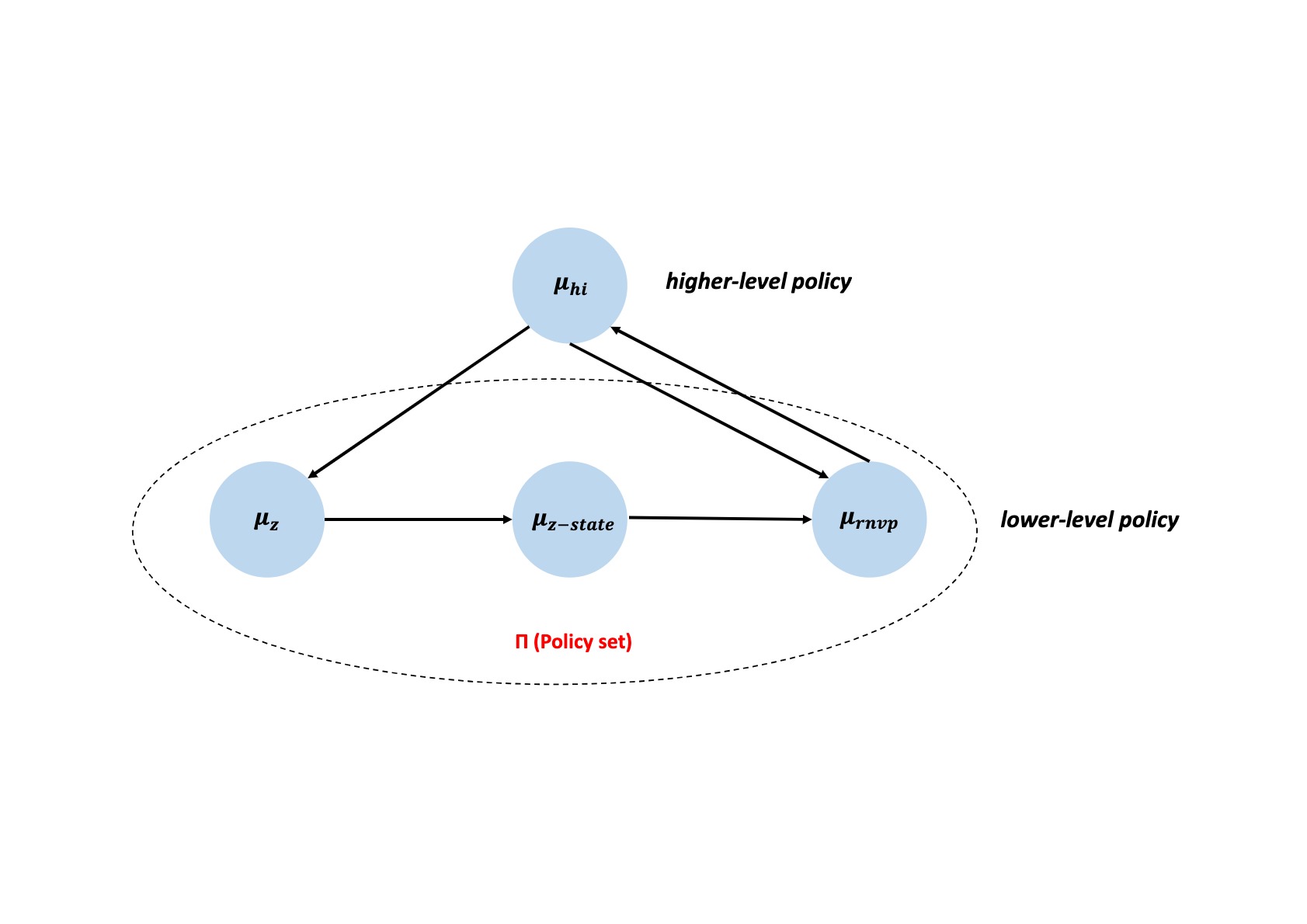}  
  \vspace{-1.8cm}
  \caption{\textcolor{black}{Communication graph representing the dependence of our model}}
  \label{fig:policy_graph}
\end{figure}

\subsection{Off-policy Correction, Convergence and Algorithm}\label{Off-policy correction}

A replay buffer collected by the higher-level policy for off-policy RL is filled with state-action-reward transitions ($s_{t}$, $g_{t}$, $a_{t,z}$, $a_{t,z-\text{state}}$, $a_{t,\text{rnvp}}$, $\Sigma{R_{t:t+c-1}}$, $s_{t+c}$) based on a higher-level transition tuple $(s_{t:t+c-1}$, $g_{t:t+c-1}$, $a_{t:t+c-1,z}$, $a_{t:t+c-1,z-\text{state}}$, $a_{t:t+c-1,\text{rnvp}}$, $R_{t:t+c-1}$, $s_{t+c})$. During off-policy training with the step \textit{t} transition of the replay buffer at time $t + c$ steps, the following inequality holds:
\begin{equation}
J_{(...,\;g_{t},...)\;\sim\;D_{R}}(\theta)_{t+c} \neq J(\theta)_{t+c}^{OPT}
\end{equation}
where $J_{(...,\;g_{t},...)\;\sim\;D_{R}}(\theta)_{t+c}$ is the objective function of $\mu_{hi}^{\theta}$ and $J(\theta)_{t+c}^{OPT}$ is the optimal objective function of $\mu_{hi}^{\theta}$ under the condition that $a_t \neq a_{t+c}{}$ as mentioned at time \textit{t + c} in Section \ref{Preliminary Knowledge:Off-policy correction}. 

In this research, \textcolor{black}{we first address the input-output dimension issue and biased log-density estimation of the FDGM through the coordinated use of $\mu_{\text{z}}$ and $\mu_{z-\text{sate}}$. Next process begins when $\mu_{\text{rnvp}}$ generates an action $a_{t+c,\text{rnvp}}$ conditioned on the outputs of $\mu_{z-\text{state}}$. These three components, which are $\mu_{\text{z}}$, $\mu_{z-\text{sate}}$ and $\mu_{\text{rnvp}}$, are then jointly trained using a shared loss function $L_{\text{common actor}}$ and $L_{\text{common critic}}$ explained in Section \ref{Experiments}.} This unified training scheme enables stable inversion of the FDGM in the lower-level policy through the following procedure:
\begin{equation}
\Tilde{g}_{t+c} \sim \mu_{\text{rnvp}}^{-1}(a_{t,\text{rnvp}}, a_{t,z-\text{state}}) 
\end{equation}
using $a_{t,\text{rnvp}}$ and $a_{t,z-\text{state}}$ taken from the replay buffer to obtain $\Tilde{g}_{t+c}$. To relabel $g_{t}$, $\Tilde{g}_{t+c}$ is utilized for the level synchronization of adjacent policies when the higher-level policy is trained after the lower-level policy is trained in a bottom-up, layerwise fashion within HRL. \textcolor{black}{As $g_{t}$ is relabelled with $\Tilde{g}_{t+c}$ through the off-policy correction of our model, it approaches $g_{t+c, ideal}$ which ideally reflects the current internal parameter $\mu_{rnvp}^{\theta}$ of $\mu_{lo}$. Consequently, the objective function of $\mu_{t+c, hi}$ converges to its optimal value, provided the agent experiences all states of the environment. This relationship is expressed as:}
\textcolor{black}{
\begin{equation}
\begin{gathered}
\text{If} \lim_{s\to\infty} \Tilde{g}_{t+c} = g_{t+c, ideal},\\\;\;\; 
\text{then} \lim_{s\to\infty} J_{(...,\Tilde{g}_{t+c},...)\sim D_{R}}(\theta)_{t+c} = J(\theta)_{t+c}^{OPT}\text{.}    
\end{gathered}
\end{equation}
}

\textcolor{black}{Finally, the goal $\Tilde{g}_{t+2c} \sim \mu_{t+2c, hi}$ represents the optimal target for both $\mu_{t+2c, z}\,and\,\mu_{\text{t+2c, rnvp}}$, as the higher-level policy $\mu_{t+2c,hi}^{\theta}$ is dynamically aligned with the current lower-level policy $\mu_{t+2c,lo}^{\theta}$.}  

Our model combines three policies of the same type, along with the common losses $L_{\text{common actor}}$ and $L_{\text{common critic}}$ which are explained in Section \ref{Experiments}, into a unified lower-level policy $\mu_{lo}$.  The dependencies between these policies are illustrated in Fig. \ref{fig:policy_graph}. Our research leverages TD3, which has been validated for its convergence \cite{9}, for all policies. As a result, the convergence of each policy level in our model depends on the interplay of the three sub-policies and the higher-level policy, adhering to the convergence characteristics of the atomic policies in off-policy training.

The dynamics of the inverse operation using our inverse model vary for each task within the environment. To address this, we utilize Algorithm \ref{alg:the_alg1} and Algorithm \ref{alg:the_alg2} to find an optimal goal for off-policy correction using our inverse model. The process is as follows:

\begin{itemize} 
\item Step 1: Set the layer dimensions (dim.) of all policies in our model and the action dimension of $\mu_{z-\text{state}}$. 
\item Step 2: Employ their respective methods to find a goal through $\mu_{lo}$. Section \ref{Experiments} provides the layer dimensions of all policies and the action dimension of $\mu_{z-\text{state}}$ based on the algorithm used in each experimental result.
\end{itemize}

\section{Experiments} \label{Experiments}
In this research, we investigate whether our inverse model in HRL, which employs direct estimation with FDGM for off-policy correction, can achieve competitive performance compared to HIRO and LSP. Our implementation is based on the HIRO open code\footnote[1]{\url{https://github.com/tensorflow/models/tree/master/research/efficient-hrl}\label{hiro_code}} and the RealNVP open code\footnote[2]{\url{https://github.com/haarnoja/sac}} from LSP, using the same optimization algorithm to ensure transparent benchmarking. As a result, we integrate our model and LSP into the HIRO code framework. We have also released our code\footnote[3]{\url{https://github.com/jangikim2/Hierarchical_Reinforcement_Learning/tree/master/distributed_inverse-efficeint-hrl-sac_210127_tfp_bijectors.bijector_dataefficient_three_reflected_888_0420}} along with detailed explanations. 

We employ a two-level hierarchical structure for HRL, as it aligns with the structure used in HIRO. The intrinsic reward function of the lower-level policy, as defined in HIRO, is used without modification. All neural networks, including the critic of the FDGM part, consist of fully connected layers, consistent with the implementation in HIRO. The only exception is the actor of the FDGM part, which is implemented as a RealNVP.

HIRO utilizes the TD3 policy gradient algorithm, a variant of DDPG designed for continuous control. Since our model is built on the HIRO source code, all policies in our model also employ TD3. Consequently, all policies within the lower-level policy set, $\Pi$, leverage the actor and critic losses of DDPG for their learning. Instead of computing separate losses for each policy in $\Pi$, the losses used for training the actor and critic of the three policies are combined into common losses, $L_{\text{common actor}}$ and $L_{\text{common critic}}$. These common losses are calculated as the average of the individual losses of all three policies, as follows:
\begin{equation} \label{loss_equation}
\begin{aligned}
L_{\text{common actor}} = ( L_{z\,actor} + L_{z-\text{state actor}} + L_{\text{rnvp actor}} ) / 3 \\
L_{\text{common critic}} = ( L_{z\,critic} + L_{z-\text{state critic}} + L_{\text{rnvp critic}} ) / 3
\end{aligned}
\end{equation}
where:

\begin{itemize}
\item $L_{z \,actor}$ and $L_{z \,critic}$ are the actor loss and critic loss of $\mu_{z}$
\item $L_{z-\text{state actor}}$ and $L_{z-\text{state critic}}$ are the actor loss and critic loss of  $\mu_{z-state}$ 
\item $L_{\text{rnvp actor}}$ and $L_{\text{rnvp critic}}$ is the actor loss and critic loss of  $\mu_{rnvp}$
\end{itemize}

The implementation of the lower-level policy and off-policy correction in the HIRO open-source code has been modified. To achieve the primary objective of the experiment, we introduce two specific modifications to the HIRO code:
\begin{enumerate}
    \item We replace the neural network of the lower-level policy in HIRO with the architecture of our model.
    
    \item We replace the indirect probabilistic method used in HIRO with the inverse operation of our model, as outlined in Algorithm \ref{alg:the_alg1} or Algorithm \ref{alg:the_alg2}, for off-policy correction. All other parts of the code remain unchanged.
\end{enumerate}    

For LSP, we replace all of its policies with the HIRO model architecture on the goal space, as designed in HIRO, except for the actor of the FDGM part, as previously mentioned. Although LSP does not originally incorporate off-policy correction, we apply off-policy correction using the inverse operation of the FDGM in the lower-level policy to ensure a fair comparison.

Since HIRO investigates the Ant environment in MuJoCo, which is part of the OpenAI Gym, the HIRO open-source code defines optimal maximum and minimum tuple values for the goals of both the higher-level and lower-level policies. In this experiment, we adopt these values as the standard goal spaces. For further details, refer to Appendix \ref{appendix:experimentdetails} and \ref{appendix-experiments}.

In our evaluation, we focus on the more challenging tasks within the Ant environment, such as Ant Push and Ant Fall, which include different modes (Multi and Single) among the tasks available in HIRO.

We conduct two comparisons between our model and HIRO, focusing on the methodology for off-policy correction and the biased log-density estimation related to FDGM:
\begin{enumerate}
    \item Optimal Goal Space Comparison in Section \ref{optimal_comparison}: We use the optimal maximum and minimum tuple values for the goals of the higher-level and lower-level policies, as defined in HIRO, as default parameters.
    
    \item **Non-Optimal Goal Space Comparison** in Section \ref{non-optimal_comparison}: We modify some dimensions of the optimal maximum and minimum tuple values for the goals of the higher-level and lower-level policies, as defined in HIRO.
\end{enumerate}

Our model is also compared with LSP in Section \ref{comparison_with_LSP} in terms of the average reward achieved by the higher-level policy, focusing on the inherent restriction between the input and output dimensions of FDGM. Additionally, an ablation study is conducted to compare a variant of our model with the original model in Section \ref{Ablation study}. Both our model and HIRO are evaluated on tasks from the MuJoCo Ant environment in OpenAI Gym, including Ant Push Multi, Ant Fall Multi, Ant Push Single, and Ant Fall Single, to assess the performance of our research. Specifically, our model is evaluated against LSP in Ant Push Single and Ant Fall Single. Further details of the experiments are provided in Appendix \ref{appendix:experimentdetails} and \ref{appendix-experiments}.

The objectives of our experiments are as follows:

\begin{enumerate}

    \item To verify that our model improves agent performance more effectively than HIRO in terms of off-policy correction.

    \item To determine the extent to which our model addresses the chronic issue of FDGM in HRL.
    
    \item To evaluate the impact of the flexible goal dimension in our model on the agent's performance compared to LSP.
    
    \item To compare our model with a variant of our model.

\end{enumerate}

\begin{figure*}
  \centering
  \subcaptionbox{Ant Push Multi\label{fig:Ant Push Multi}}{
  \centering
  \includegraphics[width=3in]{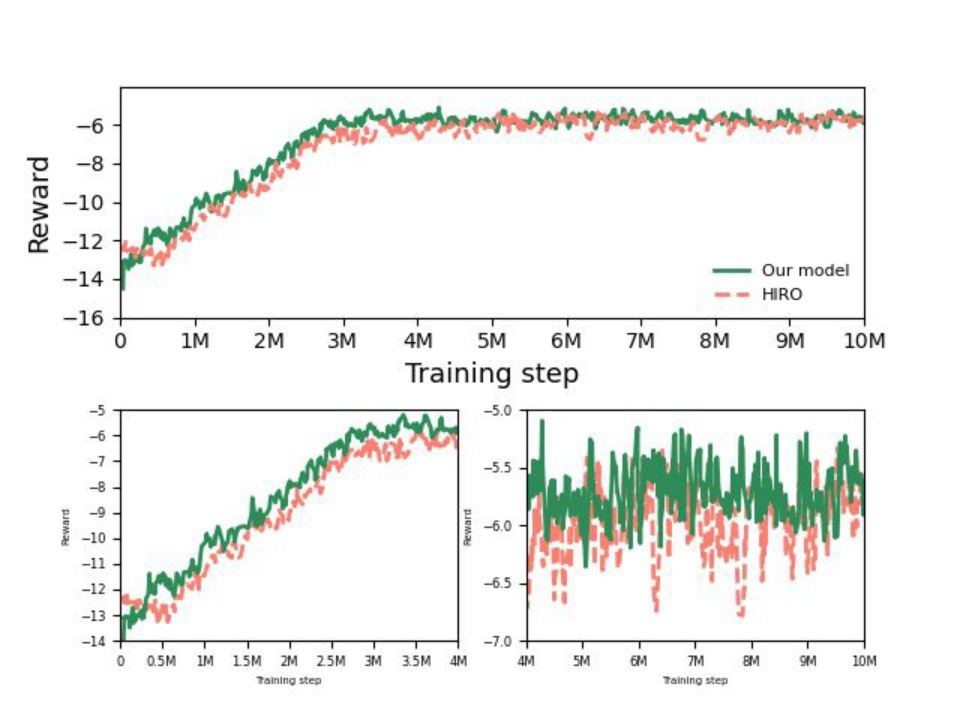}}
  \subcaptionbox{Ant Fall Multi\label{fig:Ant Fall Multi}}{
  \centering  
  \includegraphics[width=3in]{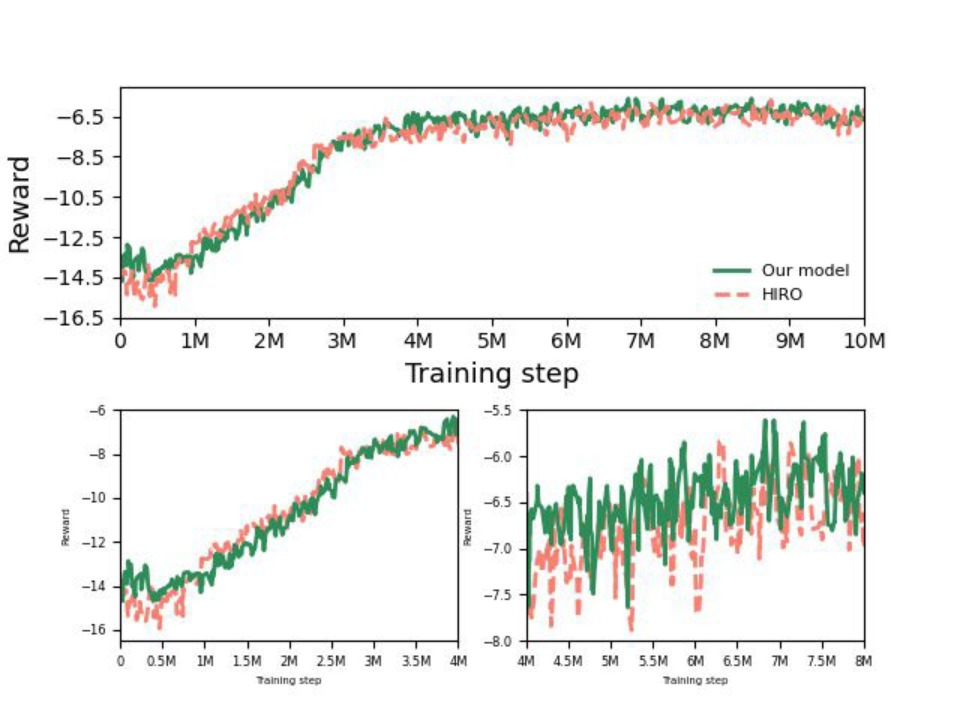}}%

  \bigskip
  \subcaptionbox{Ant Push Single\label{fig:Ant Push Single}}{
  \centering  
  \includegraphics[width=3in]{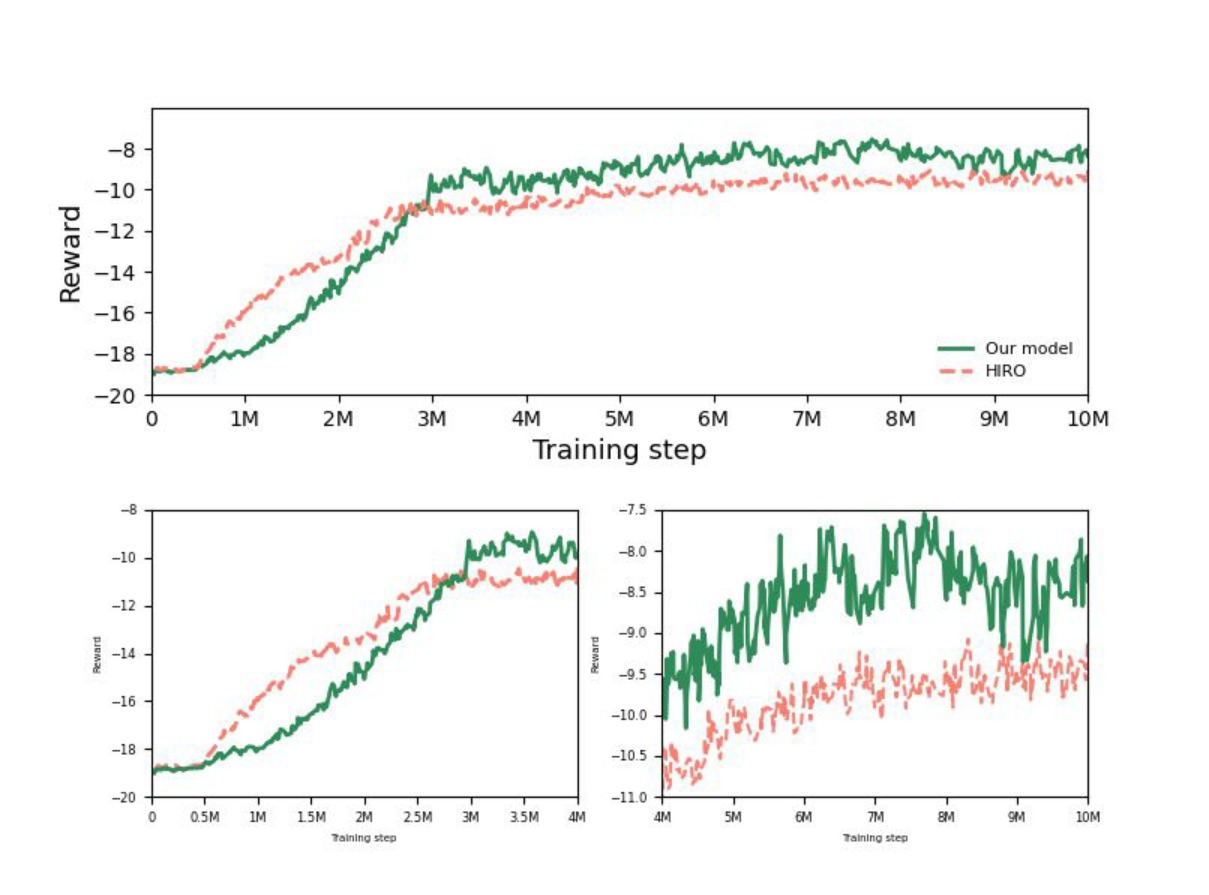}}  
  \subcaptionbox{Ant Fall Single\label{fig:Ant Fall Single}}{
  \centering  
  \includegraphics[width=3in]{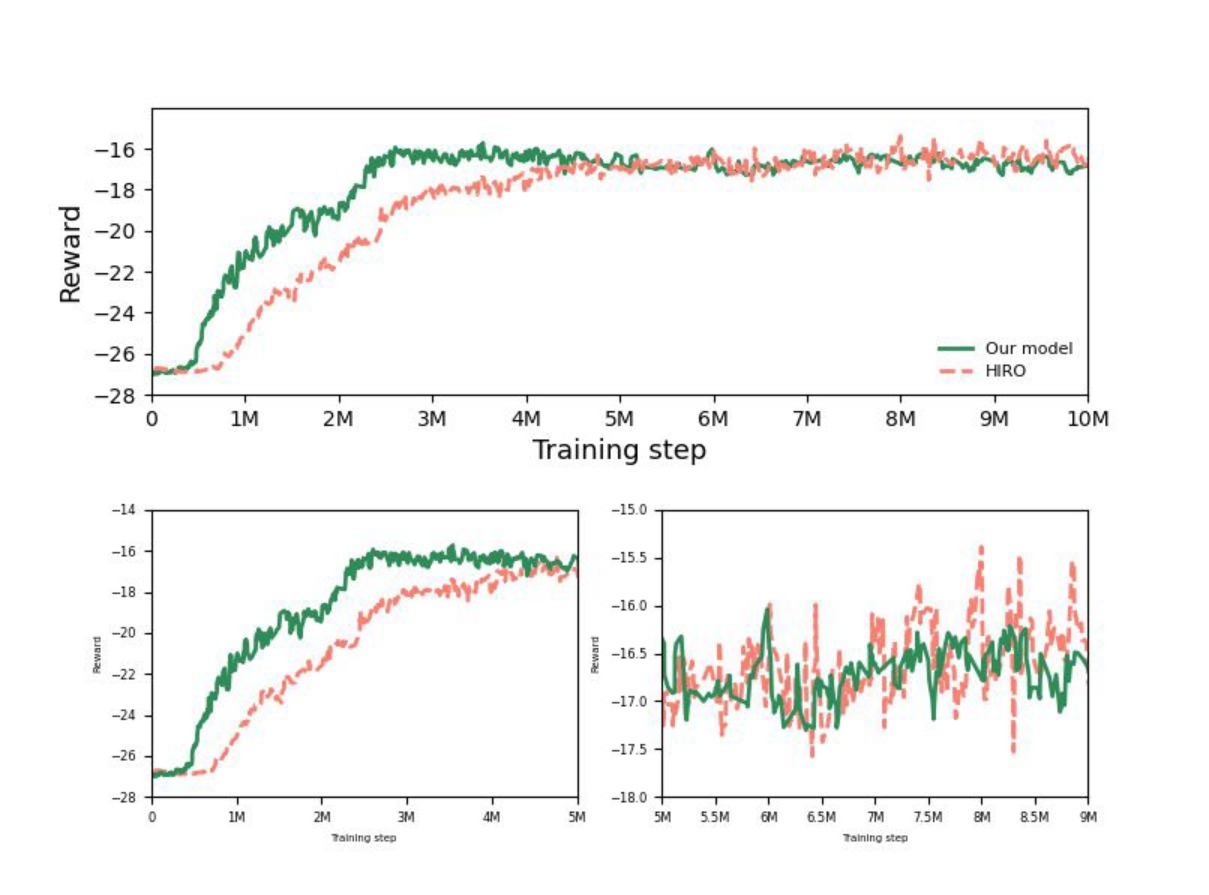}}%
  
  \caption{\textcolor{black}{For the task of the higher-level policy, the average rewards of our model (green) are} compared with that of HIRO (salmon) with the same NN size on condition of the optimal goal space.}
  \label{fig:Result_4_task}

\end{figure*}

\subsection{Comparative Analysis \textcolor{black}{with the Optimal Goal Space}}
\label{optimal_comparison}


Algorithm \ref{alg:the_alg1} and \ref{alg:the_alg2} aim to find the optimal $\Tilde{g}_{t}$ from the FDGM part, reflecting the current knowledge of the lower-level policy in each task. In Fig. \ref{fig:Result_4_task}, the average results of our model for the tasks of the higher-level policy are compared with those of HIRO using default parameters, including the optimal goal space. Our model demonstrates significantly better performance in most tasks.

\begin{itemize} 
\item Ant Push Multi --- Actor of FDGM : 145 dim., All other NNs : 170 dim., $a_{t,z-\text{state}}$ : 16 dim. --- in Fig. \ref{fig:Ant Push Multi}: Our model achieves better scores than HIRO with default parameters until 4.5M. Beyond this point, our results marginally outperform HIRO up to 10M. Algorithm \ref{alg:the_alg1} is used for our model.

\item Ant Fall Multi --- Actor of FDGM : 150 dim., All other NNs : 170 dim., $a_{t,z-\text{state}}$ : 18 dim. --- in Fig. \ref{fig:Ant Fall Multi}: Our model and HIRO with default parameters achieve similar results until 4.8M. After this, our model's performance is slightly better up to 8M. Algorithm \ref{alg:the_alg1} is used for our model.

\item Ant Push Single --- Actor of FDGM : 130 dim., All other NNs : 150 dim., $a_{t,z-\text{state}}$ : 19 dim. --- in Fig. \ref{fig:Ant Push Single}: Our model achieves considerably better performance than HIRO with default parameters until 10M, although our model's results are not as strong as HIRO's up to 3M. Algorithm \ref{alg:the_alg1} is used for our model.

\item Ant Fall Single --- All lower policy's NNs : 135 dim., All higher policy's NNs : 160 dim., $a_{t,z-\text{state}}$ : 24 dim. --- in Fig. \ref{fig:Ant Fall Single}: Our model achieves a much faster learning speed up to 5.25M compared to HIRO with default parameters. Our results are similar to HIRO's up to 8M, after which they are slightly lower until 10M. Algorithm \ref{alg:the_alg2} is used for our model.

\end{itemize}

\begin{figure*}
  \centering 
  \subcaptionbox{Ant Push Multi\label{fig:Latent Ant Push Multi}}
  {
  \centering 
  \includegraphics[width=3in]{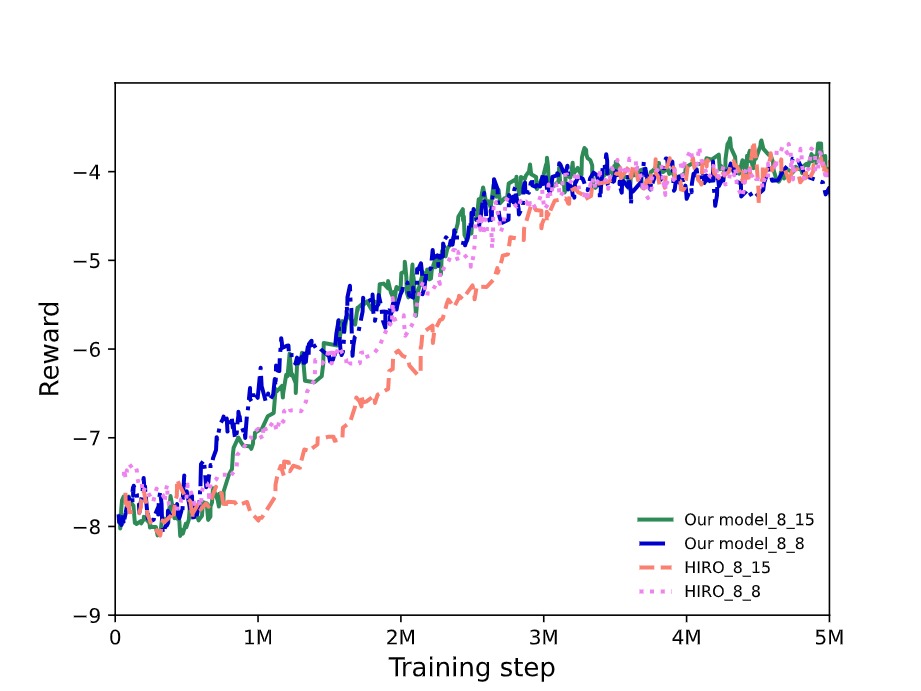}}  
  \subcaptionbox{Ant Fall Multi\label{fig:Latent Ant Fall Multi}}{
  \centering 
  \includegraphics[width=3in]{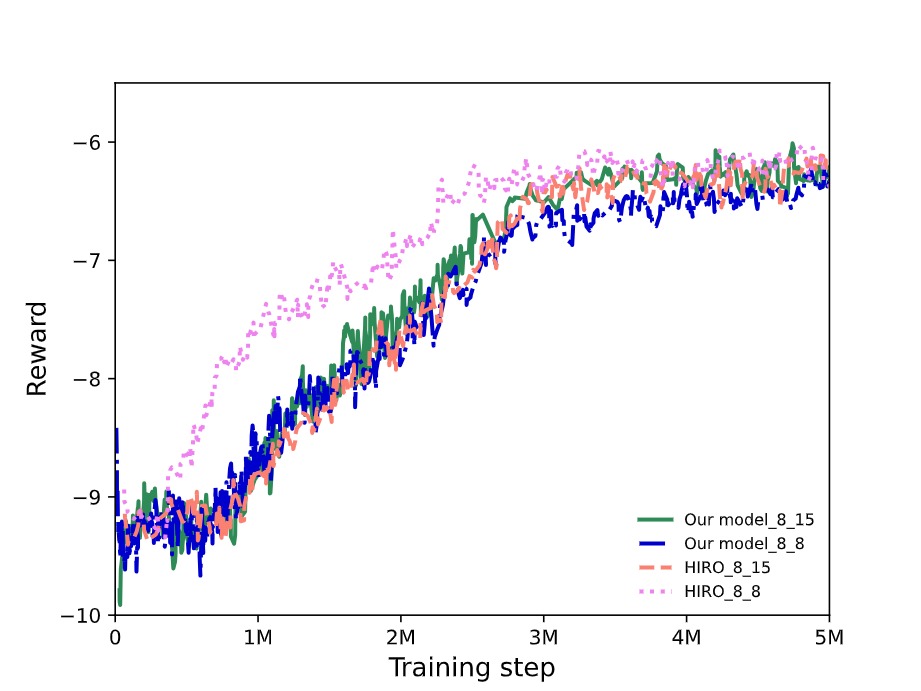}}%

  \bigskip
  \subcaptionbox{Ant Push Single\label{fig:Latent Ant Push Single}}{
  \centering 
  \includegraphics[width=3in]{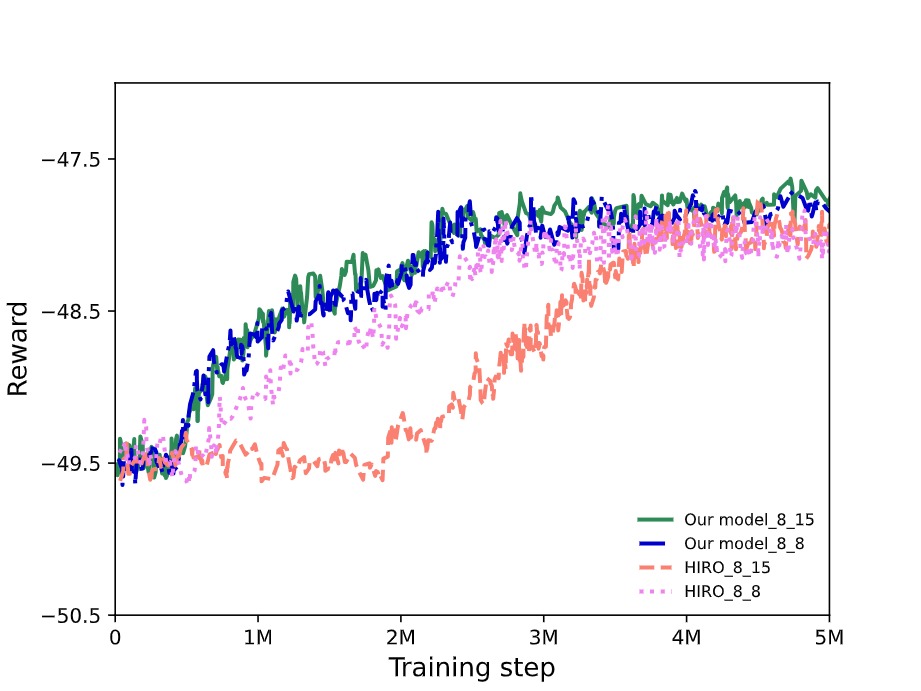}}
  \subcaptionbox{Ant Fall Single\label{fig:Latent Ant Fall Single}}{
  \centering 
  \includegraphics[width=3in]{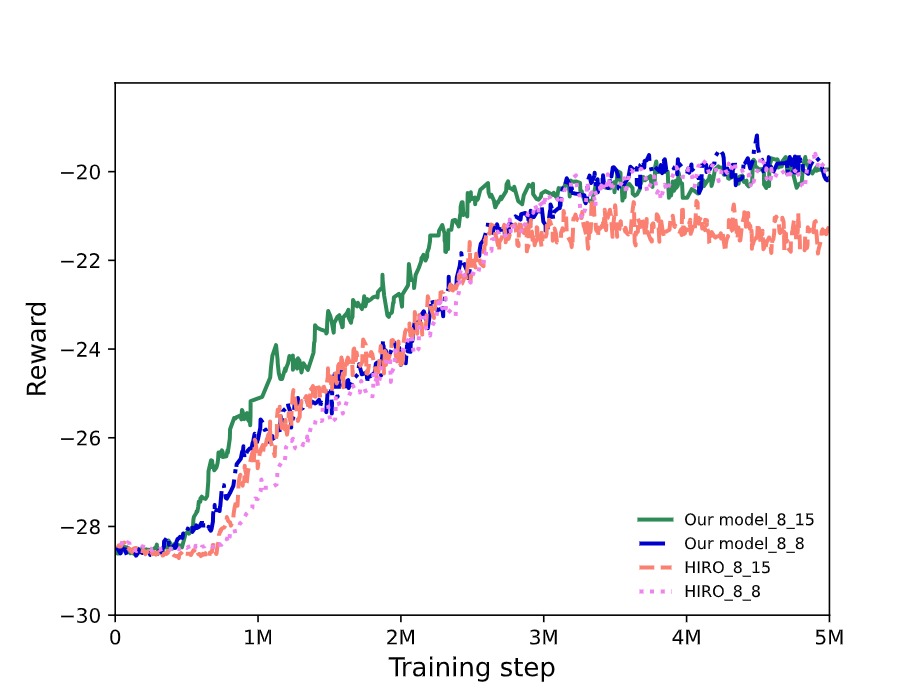}}%
  
  \caption{\textcolor{black}{The average rewards of our \textcolor{black}{model} (green, \textcolor{black}{blue}) for the task of the higher-level policy are} compared with that of HIRO (salmon, \textcolor{black}{violet}) on condition of a non-optimal goal space. \textcolor{black}{The numbers of (our model or HIRO)\textunderscore 8\textunderscore (15 or 8) are the goal dimension of higher-level policy and that of lower-level policy respectively.}}
  \label{fig:Latent Result_4_task}
\end{figure*}

\subsection{\textcolor{black}{Comparative Analysis with a Non-Optimal Goal Space}}
\label {non-optimal_comparison}

A non-optimal goal space is required for a fair comparison between the two models, HIRO and our model. We consider two cases for the non-optimal goal space:

\begin{enumerate}
    \item First Case: The higher-level policy's goal dimension is set to 8, replacing its default goal dimension of 2 or 3 depending on the task, with the first 8 goals of the lower-level policy of HIRO, whose default goal dimension is 15. This configuration is represented as (Our model or HIRO)\_8\_15, where the notation 'Model name\_goal dimension of the higher-level policy\_goal dimension of the lower-level policy' is used. As a result, the first 8 default goals of the lower-level policy of HIRO are selected as the goals for the higher-level policy of (Our model or HIRO)\_8\_15. The goal space of the lower-level policy in (Our model or HIRO)\_8\_15 remains the same as the default goal space of the lower-level policy of HIRO.

    \item Second Case: The configuration (Our model or HIRO)\_8\_8 removes the last 7 goals of the lower-level policy of (Our model or HIRO)\_8\_15, reducing the lower-level policy's goal dimension to 8.
\end{enumerate}  

Figure \ref{fig:Non_Optimal_Implementation} in the Appendix provides further details about these two goal spaces.

For the comparison with HIRO, we first tune the appropriate parameters—specifically, the layer dimensions of all policies at each level and the dimension of $a_{t,z-\text{state}}$ in the conditional part—to optimize the performance of our model in each task using Algorithm \ref{alg:the_alg1} or \ref{alg:the_alg2}. Once the best-performing parameters for our model are determined, we compare the performance of HIRO, configured with the same layer dimensions, against our model for each task.







The conditions for achieving the best performance with our model are as follows:

\begin{itemize}

\item Ant Push Multi --- Our model - Actor of FDGM : 135 dim., All other NNs : 160 dim., \textcolor{black}{Our model\textunderscore 8\textunderscore 15 - $a_{t,z-\text{state}}$ : 12 dim., Our model\textunderscore 8\textunderscore 8 - $a_{t,z-\text{state}}$ : 20 dim.} --- in Fig. \ref{fig:Latent Ant Push Multi}:  Our model\_8\_8 and Our model\_8\_15 show better performance than HIRO\_8\_8, which is the best among HIRO's two models. Algorithm \ref{alg:the_alg1} is used for our model.

\item Ant Fall Multi --- Our model - Actor of FDGM : 140 dim., All other NNs : 165 dim., \textcolor{black}{Our model\textunderscore 8\textunderscore 15 - $a_{t,z-\text{state}}$ : 12 dim., Our model\textunderscore 8\textunderscore 8 - $a_{t,z-\text{state}}$ : 15 dim.} --- in Fig. \ref{fig:Latent Ant Fall Multi}:  HIRO\_8\_8 shows the best performance among all other models. Algorithm \ref{alg:the_alg1} is used for our model.

\item Ant Push Single --- Our model - Actor of FDGM : 130 dim., All other NNs : 150 dim., \textcolor{black}{Our model\textunderscore 8\textunderscore 15 - $a_{t,z-\text{state}}$ : 9 dim., Our model\textunderscore 8\textunderscore 8 - $a_{t,z-\text{state}}$ : 22 dim.} --- in Fig. \ref{fig:Latent Ant Push Single}:  Both of our models show better performance than HIRO's two models. Algorithm \ref{alg:the_alg1} is used for our model.

\item Ant Fall Single --- Our model - Actor of FDGM : 135 dim., All other NNs : 160 dim., \textcolor{black}{Our model\textunderscore 8\textunderscore 15 - $a_{t,z-\text{state}}$ : 26 dim., Our model\textunderscore 8\textunderscore 8 - $a_{t,z-\text{state}}$ : 22 dim.} --- in Fig. \ref{fig:Latent Ant Fall Single}:  Our model\_8\_15 shows the best performance among all other models. The inverse operation of Algorithm \ref{alg:the_alg2}, combined with the NN size from Algorithm \ref{alg:the_alg1}, is used for our model.
\end{itemize}

Our model demonstrates the best performance in most tasks. However, in Fig. \textcolor{black}{\ref{fig:Latent Ant Fall Multi}}, our model's performance falls slightly behind HIRO's. Compared to tasks with a high reward shape, such as Ant Push Multi and Ant Fall Multi, our model shows significantly superior performance over HIRO in tasks with a low reward shape, such as Ant Push Single in Fig. \textcolor{black}{\ref{fig:Latent Ant Push Single}} and Ant Fall Single in Fig. \textcolor{black}{\ref{fig:Latent Ant Fall Single}}, where there is ample room for improvement. Notably, our model performs impressively in the early stages, even when the replay buffer lacks sufficient data for off-policy correction. This is primarily because our model can leverage an exact goal, $\Tilde{g_{t}}$, through direct inference from the inverse operation of FDGM.

\begin{figure*}
  \centering 
  \subcaptionbox{Ant Push Single\label{fig:Latent Ant Push Single 15}}{
  \centering 
  \includegraphics[width=3in]{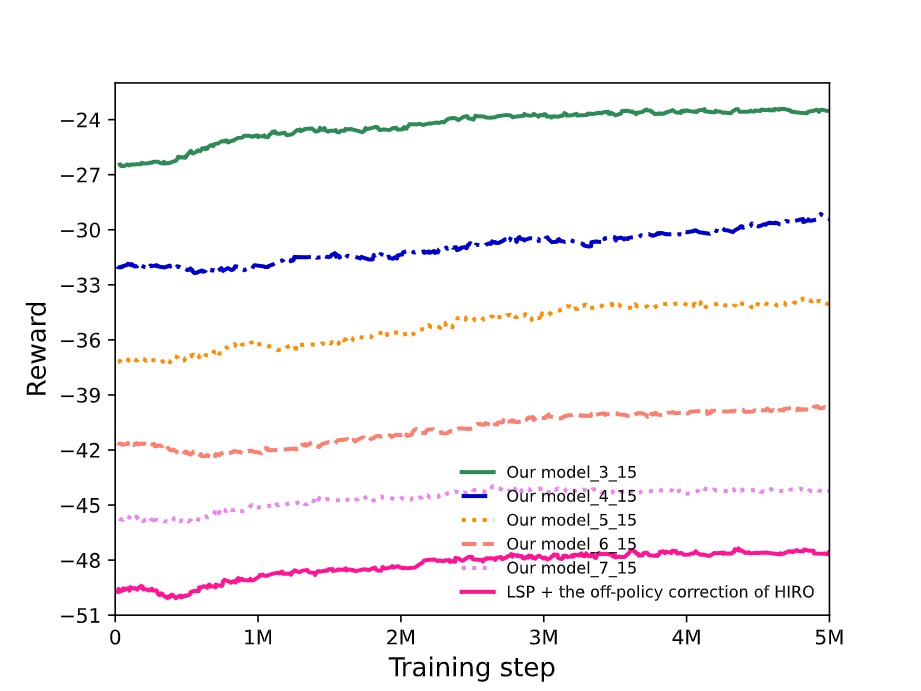}}
  \subcaptionbox{Ant Push Single\label{fig:Latent Ant Push Single 8}}{
  \centering 
  \includegraphics[width=3in]{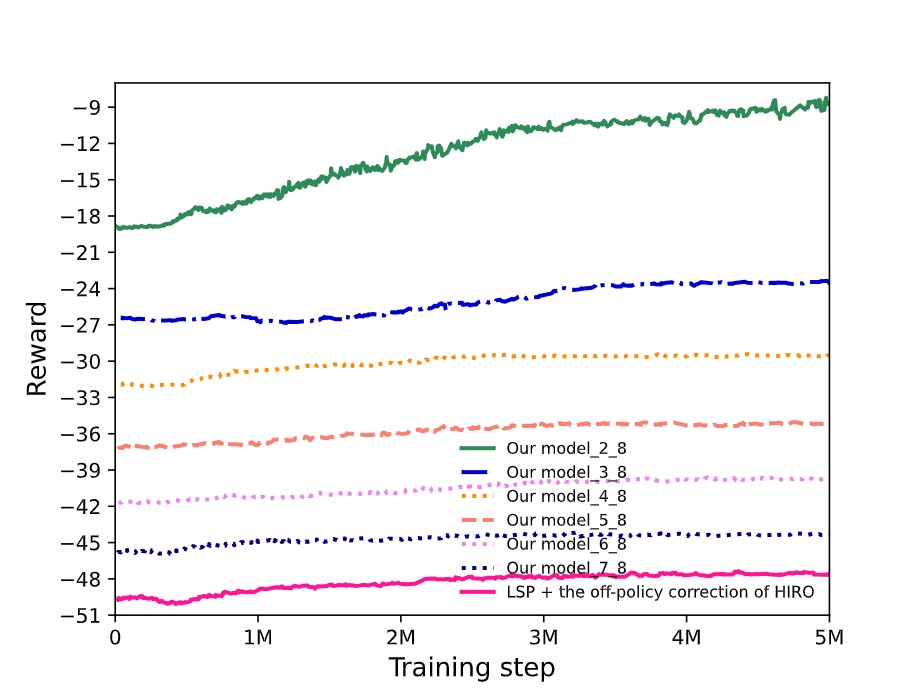}}%

  \bigskip
  \subcaptionbox{Ant Fall Single\label{fig:Latent Ant Fall Single 15}}{
  \centering 
  \includegraphics[width=3in]{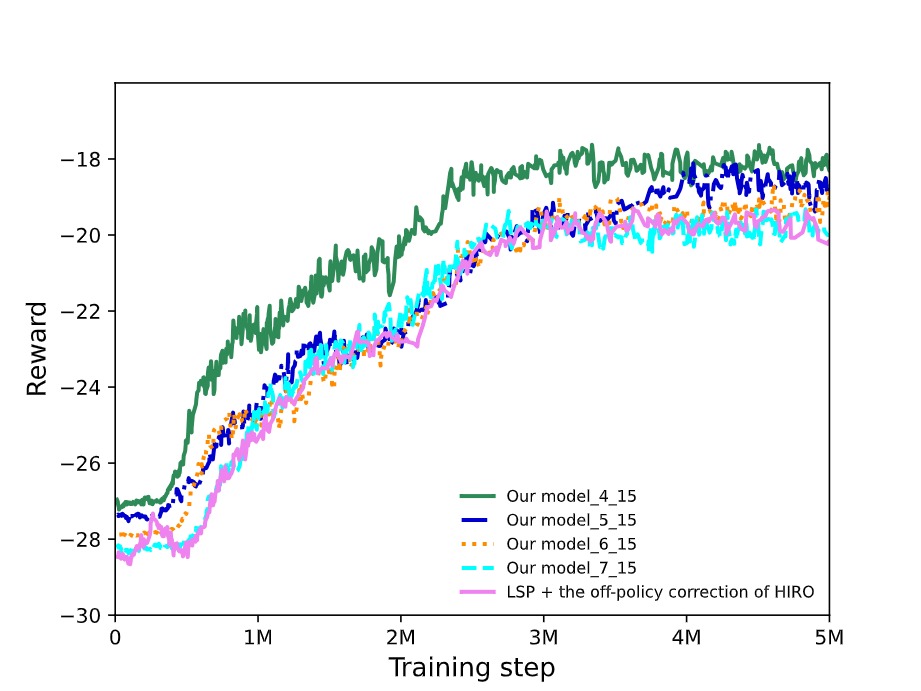}}
  \subcaptionbox{Ant Fall Single\label{fig:Latent Ant Fall Single 8}}{
  \centering
  \includegraphics[width=3in]{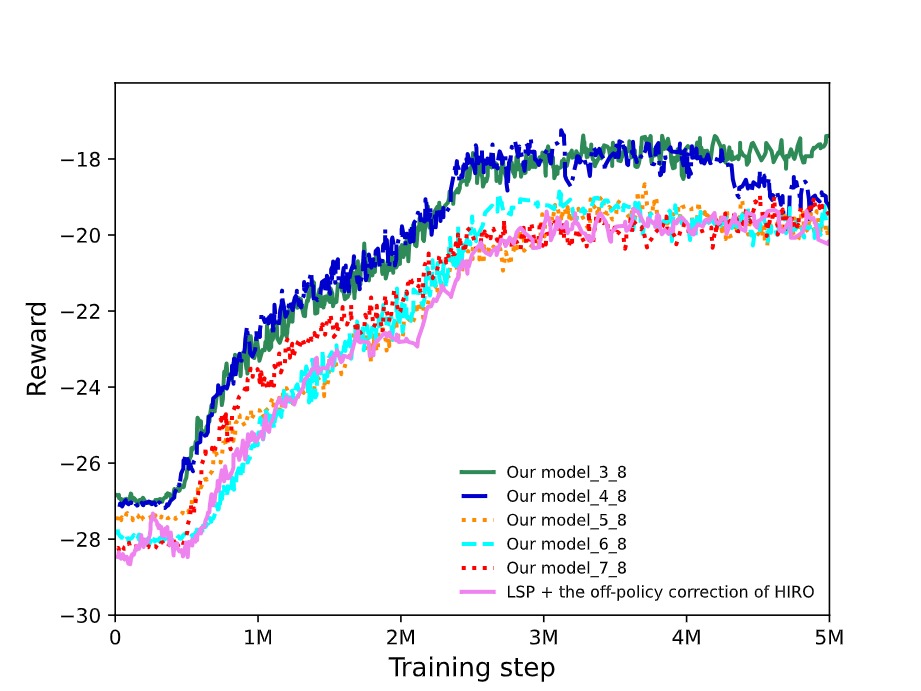}}%

  \caption{\textcolor{black}{For the task of the higher-level policy, the average rewards of our model (green, blue, orange, salmon, violet and navy for Ant Push Single and green, blue, orange, cyan and red for Ant Fall Single) are compared with that of LSP (pink for Ant Push Single and violet for Ant Fall Single). They are based on the change of higher-level goal dimension of a) our model\textunderscore 8\textunderscore 15}, (b) our model\textunderscore 8\textunderscore 8 (c) our model\textunderscore 8\textunderscore 15 (d) our model\textunderscore 8\textunderscore 8 used in the corresponding task of Fig. \ref{fig:Latent Result_4_task}.}
  \label{fig:Latent Result_4_task_15_8}
\end{figure*} 

\subsection{Comparative Analysis Between Our Model and LSP}
\label{comparison_with_LSP}

One of the primary objectives of our model is to enable flexible goal spaces for all policies in HRL without any restrictions. Our model\textunderscore(2,3,4,5,6 or 7)\textunderscore (8 or 15) exemplifies this goal flexibility. These non-optimal goal spaces are constructed using the same method as (Our model or HIRO)\_8\_(8 or 15). Specifically, the goal space of the higher-level policy is replaced with the first x goals (where x = 2,3,4,5,6 or 7) of the lower-level policy of HIRO, whose default goal dimension is 15. Fig. \ref{fig:Non_Optimal_Implementation} in the Appendix provides examples of these goal spaces.

In Fig. \ref{fig:Latent Result_4_task_15_8}, our model\textunderscore(2,3,4,5,6 or 7)\textunderscore (8 or 15) is compared with LSP to demonstrate the superiority of our model's goal flexibility. The layer dimensions of our model are the same as those in the corresponding tasks of our model\_8\_(8 or 15) in Fig. \ref{fig:Latent Result_4_task}.

LSP, due to the invertible bijective transformation property of FDGM, requires that the output dimension for selecting a goal space for all policies in HRL must match the action space dimension of the environment, which is 8 in the MuJoCo Ant environment. As a result, LSP is constrained to LSP\_8\_8. The layer dimensions of LSP are configured based on the settings of our model in Ant Push Single in Fig. \textcolor{black}{\ref{fig:Latent Ant Push Single}} and Ant Fall Single in Fig. \textcolor{black}{\ref{fig:Latent Ant Fall Single}}. Additionally, the observation embedding dimension of LSP is tuned to achieve optimal performance, as follows:

\begin{itemize}
\item Ant Push Single --- Higher policy's NN : 150 dim., Lower policy's NN : 130 dim., Output dimension of each level's observation embedding : 23 dim. --- in Fig. \ref{fig:Latent Ant Push Single 15} and \ref{fig:Latent Ant Push Single 8}

\item Ant Fall Single --- Higher policy's NN : 160 dim., Lower policy's NN : 135 dim., Output dimension of each level's observation embedding : 14 dim. --- in Fig. \ref{fig:Latent Ant Fall Single 15} and \ref{fig:Latent Ant Fall Single 8}
\end{itemize}

Fig. \ref{fig:Latent Result_4_task_15_8} demonstrates the effectiveness of selecting the goal dimension for the higher-level policy of our model. It reveals that smaller goal dimensions for the higher-level policy of our model lead to better performance. This trend suggests that the goal dimension of the higher-level policy of our model approaches the default goal dimension of HIRO's higher-level policy. LSP performs significantly worse than our model, underscoring the superiority of our model's flexibility in goal dimension.

\subsection{Ablation Study}
\label{Ablation study}

\begin{figure}
 \centering
 \includegraphics[scale=0.25]{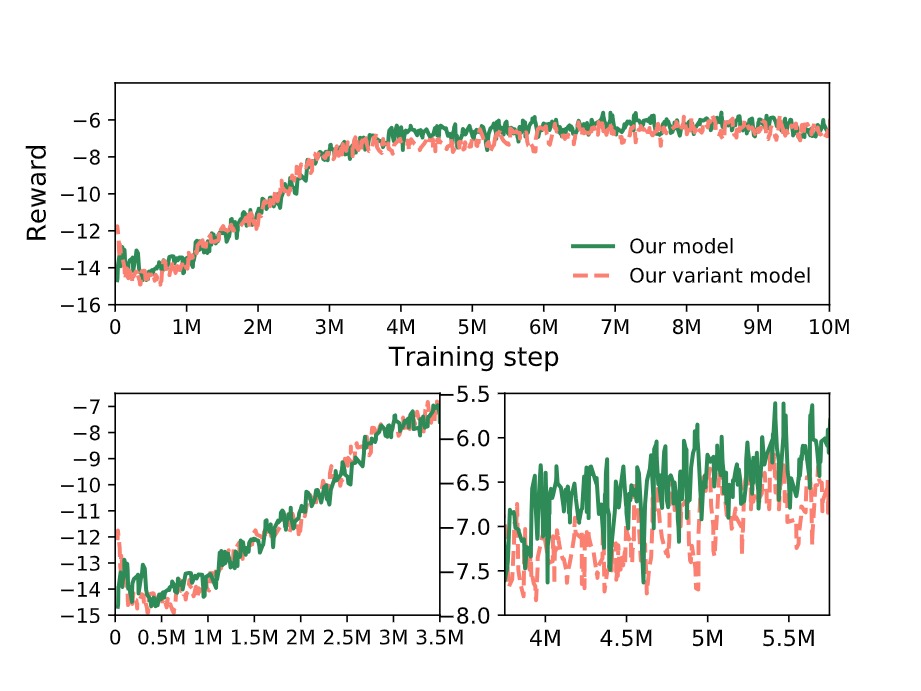} 
 \caption{Average result of our method (green) compared with that of the variant model (salmon) in Ant Fall Multi}
 \label{fig:variant_result}
\end{figure}

The importance of $a_{t,z}$ is of particular interest due to its significant role in addressing the chronic issue of FDGM, specifically the biased log-density problem. To explore its impact on our model's performance, we investigate how replacing $a_{t,z}$ with the goal $g_{t}$ as the input to the conditional part affects the model's performance.

\begin{itemize}
\item Ant Fall Multi of variant model -- Actor of FDGM : 150, All other NNs : 180, $a_{t,z-\text{state}}$ = 19
\end{itemize}

In this experiment, we do not set the parameters of each policy to match those of our model in the comparative analysis. Instead, we first identify the parameters of the variant model with the optimal goal space to achieve its best performance in the Ant Fall Multi task from Section \ref{optimal_comparison}. We then compare the performance of our original model from the previous Ant Fall Multi experiment with the optimal goal space to the variant model with the newly determined parameters.

The results show that while both models perform similarly until 3.5M, our original model marginally outperforms the variant model up to 7.5M. After this point, both models exhibit similar performance up to 10M. The performance comparison between our model and the variant model is illustrated in Fig. \ref{fig:variant_result}.

\section{Discussion} \label{Discussion}
\subsection{The Performance of the Higher-Level Policy Based on Our Model in All Tasks}
In most experiments, our model outperforms both HIRO and LSP. Despite not relying on indirect estimation methods, such as the indirect probabilistic approach used in HIRO, our model achieves competitive performance against HIRO and LSP by leveraging the inverse operation of FDGM. Notably, the lower-level policy, which consists of three sub-policies, does not significantly hinder the performance of the higher-level policy during the early learning stages. However, fine-tuning the dimension of $a_{t,z-\text{state}}$ and the layer dimensions of both level policies is essential, as these parameters influence the performance of the higher-level policy.

\subsection{\textcolor{black}{The Inference from the Performance of Our Model Compared with LSP}}
Even with a non-optimal goal space, our model demonstrates competitive performance, underscoring the importance of flexibility in choosing the goal dimension in HRL. In Fig. \ref{fig:Latent Result_4_task_15_8}, LSP fails to achieve competitive performance due to its inherent limitation of a fixed goal space. In contrast, our model showcases exceptional performance potential, thanks to its flexibility in selecting the goal space.

\subsection{\textcolor{black}{Why is RealNVP Selected as a FDGM in Our Model?}}
\textcolor{black}{A persistent challenge in FDGMs is biased log-density estimation, which remains an active area of research  \cite{56, 57, 52, 53, 54}. Although RealNVP represents an early research of FDGMs, it has become a standard benchmark in the field due to its foundational contributions. Consequently, RealNVP serves as an appropriate baseline model for our model.
}

\subsection{\textcolor{black}{Can Our Model Function as an Inverse Model in HRL?}}
\textcolor{black}{Our model demonstrates valid inverse modeling capabilities in HRL, as the lower-level policy $\mu_{lo}$ admits an exact invertible representation through three internal components which are $\mu_{\text{z}}$, $\mu_{z-\text{sate}}$ and $\mu_{\text{rnvp}}$.
}

\subsection{\textcolor{black}{Is the Role of $\mu_{z-\text{sate}}$ Important for Our Model?}}
\textcolor{black}{The performance of our model is sensitive to the role of $\mu_{z-\text{sate}}$ supporting an optimized conditional input to $\mu_{\text{rnvp}}$. $\mu_{z-\text{sate}}$ decides the optimized synchronization between the global representation of $\mu_{\text{z}}$ and the local representation of $\mu_{\text{rnvp}}$.}

\subsection{\textcolor{black}{Is There Any Research in Terms of Off-Policy Correction for HRL Since HIRO?}}
\textcolor{black}{There is no research regarding off-policy correction for HRL since HIRO. Although we have investigated a relevant research \cite{58, 59, 60}, all researches  are different from the off-policy correction of model-free on-line HRL.}

\subsection{\textcolor{black}{The Inverse Operation and Layer Dim. in Algorithm \ref{alg:the_alg1} and Algorithm \ref{alg:the_alg2}}}

The rationale for considering Algorithm \ref{alg:the_alg1} and \ref{alg:the_alg2} is to address the performance level and speed of the higher-level policy, respectively. Since our model employs TD3 for its policies, the FDGM part also incorporates a combination of actor and critic networks. Interestingly, when the layer dimensions of the actor match those of all other neural networks in Algorithm \ref{alg:the_alg1}, the performance of the higher-level policy falls short of expectations. However, when there is a difference in the layer dimensions, particularly for the actor of FDGM compared to the other neural networks, the performance aligns with our expectations.

Algorithm \ref{alg:the_alg1} performs a straightforward inverse operation that reflects only the FDGM part of the lower-level policy. In contrast, Algorithm \ref{alg:the_alg2} considers all three sub-parts of the lower-level policy, incorporating different layer dimensions between the higher-level and lower-level policies. Through this research, we observe that the performance level of Algorithm \ref{alg:the_alg1} is superior to that of Algorithm \ref{alg:the_alg2}. However, Algorithm \ref{alg:the_alg2} tends to achieve faster performance. In one of our experiments with a non-optimal goal space—specifically, Ant Fall Single—we combine the two algorithms to leverage their respective strengths.

\subsection{The Performance Difference Between the Low Reward Shape and the High Reward Shape}
Although our model achieves better performance than HIRO across all tasks, there is a noticeable difference in performance between high-reward-shape tasks (Ant Push Multi and Ant Fall Multi) and low-reward-shape tasks (Ant Push Single and Ant Fall Single). This discrepancy may be attributed to two main reasons. First, the chronic issue of FDGM—biased log-density estimation—still affects the inverse operation of FDGM. Specifically, the RealNVP used in our implementation is an early version of FDGM, which may contribute to this limitation. Second, the reconstruction loss of the Autoencoder also plays a significant role. In our model, the forward part and conditional part function as an encoder and decoder, respectively. In environments where the size of the feature space is much larger than that of the action space, the reconstruction loss of the Autoencoder leads to varying performance in tasks with low or high reward shapes.

\subsection{\textcolor{black}{The Further Research for an Environment with a Small Action Space and a Big Feature Space}}

The inputs to the conditional part are \(s_{t}\) and \(a_{t,z}\). In environments where the action space is small and the feature space is large, the model architecture of \cite{39} for unbiased log-density estimation in our model may not function effectively. This is because the dimension of the global representation, \(a_{t,z}\), is significantly smaller than that of \(s_{t}\). To address this, a decoder for \(a_{t,z}\) is required to increase the dimension of the global representation.

\begin{figure*}
  \vspace{-0.2cm}
  \centering
  \hspace*{-0.5cm} 
  \includegraphics[scale=0.25]{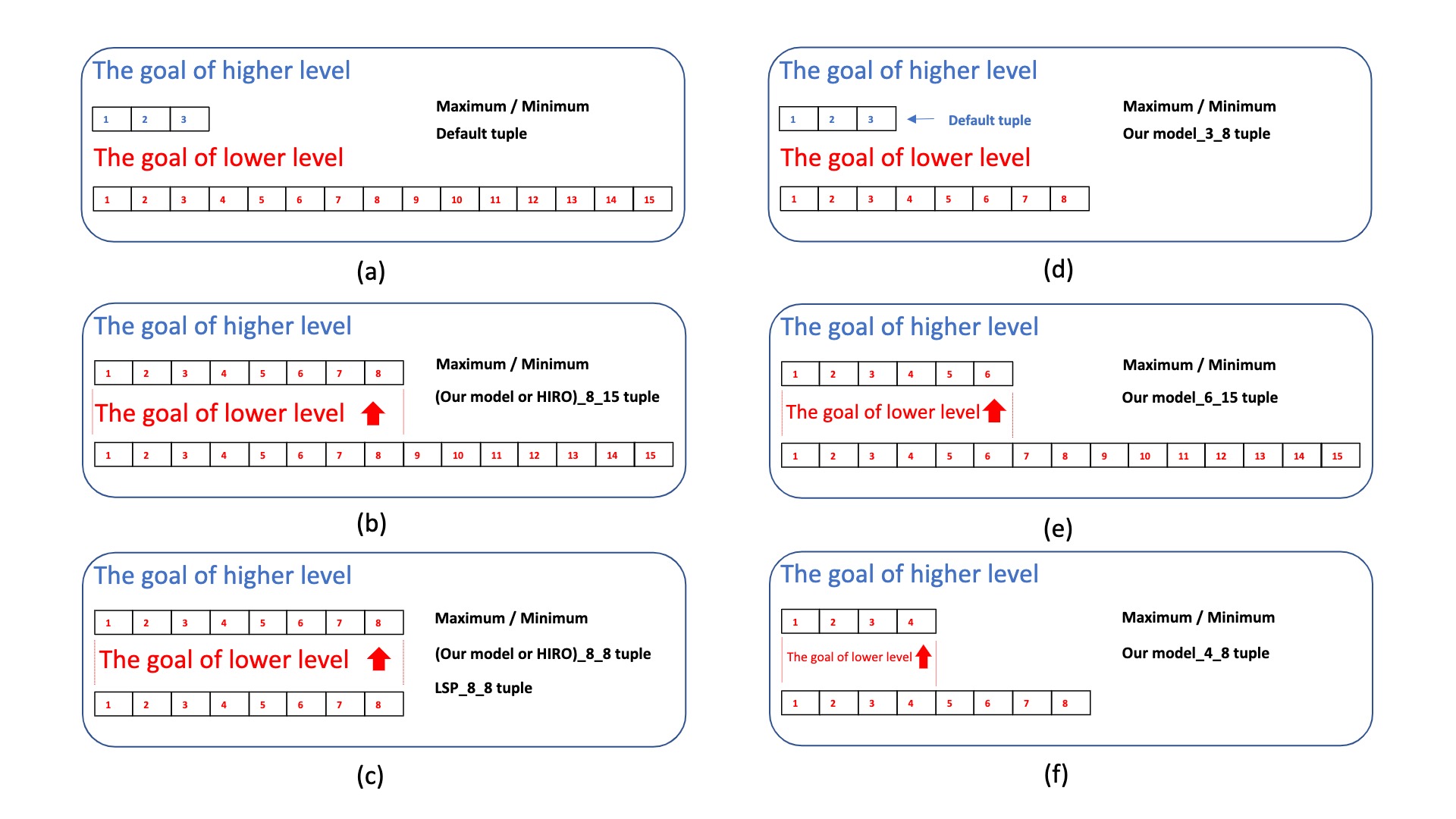}  
  \vspace{-0.5cm}
  \caption{An example of how to create a non-optimal goal space for the higher-level and lower-level policies of our model, starting from the default optimal goal space tuples of HIRO's higher-level and lower-level policies. Each level's default tuple in (a) is represented with a color—blue for the higher-level policy or red for the lower-level policy—and the position number of the tuple element. Only the tuple elements of the destination which is higher-level's goal are replaced by the corresponding tuple elements of the source which is lower-level's goal. In the case of (Our model, HIRO, or LSP)\_8\_8 for the lower-level's goal, the last 7 tuple elements of the lower-level goal are removed. (a) The default tuple, which is the optimal tuple, for the higher-level and lower-level policies of HIRO. Appendix \ref{appendix:experimentdetails} provides the definition of each level's tuple. (b) Our model\textunderscore 8\textunderscore 15 (c) Our model\textunderscore 8\textunderscore 8 (d) Our model\textunderscore 3\textunderscore 8:  The goal space of the higher-level policy for 'Our model\_2\_8' and 'Our model\_3\_8' is constrained to match the default goal space of Ant Push Single and Ant Fall Single in HIRO, respectively. (d) Our model\textunderscore 6\textunderscore 15 (e) Our model\textunderscore 4\textunderscore 8}
  \label{fig:Non_Optimal_Implementation}
\end{figure*}

\section{Conclusion} \label{Conclusion}
We have introduced a novel hierarchical reinforcement learning (HRL) inverse model that supports level synchronization using FDGM for off-policy correction in HRL. Our model outperforms HIRO and LSP, as demonstrated in four test tasks within the Ant environment, which includes two reward shapes, high reward shape and low reward shape, in a two-level hierarchical structure. Our model presents a generalizable approach for the inverse model in model-free HRL.

Our research demonstrates that off-policy correction in HRL can be effectively implemented using data acquired from a policy based on FDGM. This approach, known as the direct method, is more practical and broadly applicable, as it relies on natural data computation without requiring peripheral information used in indirect estimations.

To date, there has been limited research on FDGM in the RL domain. Typically, most RL research utilizes general neural networks, such as fully connected neural networks. However, as RL may require additional functionality, such as inverse operations, future RL research could benefit significantly from incorporating generative models.

Further research is needed to enhance our model. Future studies should focus on RL using general FDGM algorithms with feature-based data, rather than image-based data, which is large in size. Additionally, our generalized inverse model could be explored in atomic RL, even though it has primarily been studied in HRL. We also consider extending our work to non-Markovian environments in the future.

\printbibliography

\section{Appendices}


\appendix

Fig.~\ref{fig:Non_Optimal_Implementation} illustrates an example of creating a non-optimal goal space for the higher-level and lower-level policies of our model, starting from the default optimal goal space tuples of the higher-level and lower-level policies in HIRO.

Additionally, the training parameters for our model and the key points for the experiments are summarized as follows.

\section{The Training Parameters for Our Model}
\label{appendix:experimentdetails}

The training parameters defined in our model for all tasks are the same as those of HIRO, as follows:

\begin{itemize} 
\label{real-hyper-parameter}
\item Discount $\gamma$ = 0.99 for both controllers. 
\item Adam optimizer; actor learning rate 0.0001; critic learning rate 0.001.
\item  Soft update targets $\tau$ = 0.005 for both controllers.
\item Replay buffer of size 200,000 for both controllers. 
\item Lower-level train step and target update performed every 1 environment step.
\item Higher-level train step and target update performed every 10 environment steps.
\item No gradient clipping. 
\item Reward scaling of 1.0 for lower-level; 0.1 for higher-level.
\item Lower-level exploration is Gaussian noise with $\sigma$ = 1.0.
\item Higher-level exploration is Gaussian noise with $\sigma$ = 1.0.
\item  state $s_{t}$ : 30 dim.
\item  action $a_{t,z}$ : 8 dim.
\item Training step : 10M steps
\item Higher-level goal space for ant\_fall\_(multi or single) \\- meta\_context\_range = ((-4, -4, 0), (12, 28, 5))
\\- goal dim. in higher-level policy : 3 dim.
\item Higher-level goal space for ant\_push\_(multi or single) \\- meta\_context\_range = ((-16, -4), (16, 20))
\\- goal dim. in higher-level policy : 2 dim.
\item lower-level goal space : 
\\- CONTEXT\_RANGE\_MIN = (-10, -10, -0.5, -1, -1, -1, -1, -0.5, -0.3, -0.5, -0.3, -0.5, -0.3, -0.5, -0.3)
\\- CONTEXT\_RANGE\_MAX = ( 10,  10,  0.5,  1,  1,  1,  1,  0.5,  0.3,  0.5,  0.3,  0.5,  0.3,  0.5,  0.3)
\\- goal dim. in lower-level policy : 15 dim.
\end{itemize}

\section{The Key Points for Experiments}
\label{appendix-experiments}
\textcolor{black}{The key points to note during the experiment are reiterated below:} 


\begin{itemize} 

\item Although the dimensions of $a_{z}$ and $a_{rnvp}$ are fixed based on the environment and the input of FDGM, the dimension of $a_{t,z-\text{state}}$ is determined through a trade-off process involving experimentation to identify the optimal dimension.

\item The layer dimensions of all policies of 'Our model\textunderscore 8\textunderscore 8' in Fig. \ref{fig:Latent Result_4_task} are the same as those in 'Our model\textunderscore 8\textunderscore 15'.

\item In Fig. \ref{fig:Latent Result_4_task_15_8}, the goal space of the higher-level policy for 'Our model\_2\_8' and 'Our model\_3\_8' is constrained to match the original goal space of Ant Push Single and Ant Fall Single in HIRO, respectively.

\item The layer dimensions of all policies in our model in Fig. \ref{fig:Latent Result_4_task_15_8} are the same as those in our model in Fig. \ref{fig:Latent Result_4_task}.

\item The optimal size of \( a_{t,z-\text{state}} \) for each experiment in Fig. \ref{fig:Latent Result_4_task_15_8} is selected through experimentation.

\item Similar to Ant Fall Single in Fig. \ref{fig:Latent Result_4_task}, our model in Ant Fall Single in Fig. \ref{fig:Latent Result_4_task_15_8} leverages the combination of the inverse operation from Algorithm \ref{alg:the_alg2} and the neural network size from Algorithm \ref{alg:the_alg1}.

\item Although the LSP framework does not inherently include off-policy correction, LSP in Fig. \ref{fig:Latent Result_4_task_15_8} incorporates off-policy correction to ensure a fair comparison in the experiment.

\end{itemize}

\section{The Comparison Table}
\label{appendix-comparison-table}

\textcolor{black}{
Our proposed model and all reference models are evaluated based on two key criteria: (1) off-policy correction capability and (2) FDGM integration. In Table \ref{Comparison_model}, 'O' indicates that a model satisfies a given criterion, while 'X' denotes its absence.
}

\begin{table}[H]
    \caption{Model comparison}
    \begin{minipage}{\columnwidth}
    \begin{center}
    \begin{tabular}{ | c | c | c |}
      \hline
      \thead{ } & \makecell{Off-policy \\ correction} & \makecell{FDGM} \\
      \hline
      \thead{Our\\model} &  \makecell{O}  & O  \\
      \thead{HIRO} &  \makecell{O}  & X  \\
      \thead{LSP} &  \makecell{X}  & O  \\
      \thead{Ma et al.} &  \makecell{X}  & O  \\
      \hline
    \end{tabular}
    \end{center}
    \end{minipage}
    \label{Comparison_model}
\end{table}


\end{document}